# Evaluating explainable artificial intelligence methods for multi-label deep learning classification tasks in remote sensing

**Ioannis Kakogeorgiou**[a, *] **and Konstantinos Karantzalos**[a, b]

[a] Remote Sensing Laboratory, National Technical University of Athens, Zographou, 15780, Greece; karank@central.ntua.gr

[b] Athena Research Center, Athens, Greece

**\*** Correspondence: gkakogeorgiou@central.ntua.gr; Tel.: +302107721673

**Abstract:** Although deep neural networks hold the state-of-the-art in several remote sensing tasks, their black-box operation hinders the understanding of their decisions, concealing any bias and other shortcomings in datasets and model performance. To this end, we have applied explainable artificial intelligence (XAI) methods in remote sensing multi-label classification tasks towards producing human-interpretable explanations and improve transparency. In particular, we utilized and trained deep learning models with state-of-the-art performance in the benchmark BigEarthNet and SEN12MS datasets. Ten XAI methods were employed towards understanding and interpreting models' predictions, along with quantitative metrics to assess and compare their performance. Numerous experiments were performed to assess the overall performance of XAI methods for straightforward prediction cases, competing multiple labels, as well as misclassification cases. According to our findings, *Occlusion, Grad-CAM* and *Lime* were the most interpretable and reliable XAI methods. However, none delivers high-resolution outputs, while apart from *Grad-CAM*, both *Lime* and *Occlusion* are computationally expensive. We also highlight different aspects of XAI performance and elaborate with insights on black-box decisions in order to improve transparency, understand their behavior and reveal, as well, datasets' particularities.

## 1. Introduction

Deep neural networks have achieved remarkable success in real-world applications in various engineering fields (Deng and Yu, 2014) as well as in remote sensing (RS) (Ma et al., 2019), (Cheng et al., 2018), (Zhang et al., 2020), (Hamylton et al., 2020). However, from a scientific standpoint, black-box artificial intelligence (AI) solutions with non or questionable transparency, interpretability, and explainability are still barriers. Contrary to more simple and self-explaining models (e.g., linear regression), deep neural networks lack interpretability due to their non-linear and complex design. Even though deeper models can identify and model complex patterns as well as enable significantly higher performance, the deployment of black-box solutions in remote sensing and other disciplines is not straightforward for critical decision making (Camps-Valls et al., 2020).

To tackle this challenge, explainable AI (XAI) methods could provide human interpretable explanations to better understand machine learning black-box decisions. XAI methods could help users/ practitioners further evaluate their models beyond standard performance metrics (e.g., accuracy metric) by analyzing and inspecting individual predictions through examining their explanations (Ribeiro et al., 2016). Moreover, these methods could potentially reveal biases in the trained dataset, classes, multiple labels, and other spurious or artifactual correlations learned by a model (Lapuschkin et al., 2019). Moreover, further insights could be gained, in cases, e.g., that a model surpasses human performance; it may have encompassed scientific knowledge that can be extracted via an XAI method providing insights to the domain experts and scientific community (Samek et al., 2021).

The majority of RS studies that have considered XAI methods have been conducted for applications in the area of bio- and geosciences (Roscher et al., 2020). Regarding multispectral and radar satellite data, *Regression Activation Maps* (*RAM*) were integrated into the pipeline of (Wolanin et al., 2020) methodology providing insight into the relevant conditions leading to crop yield variability using MODIS imagery. Additionally, (Yessou et al., 2020) utilized

*Layer-wise Relevance Propagation* (*LRP*) (Bach et al., 2015) heatmaps to explain the characteristics of the different loss functions that have been examined. Recurrent Neural Networks for land use classification (Campos-Taberner et al., 2020) and crop yield estimation (Pérez-Suay et al., 2020) were scrutinized, looking at the hidden units distribution. (Levering et al., 2020) developed an interpretable by-design CNN model in order to understand the connections between landscape scenicness and the presence of landcover classes. So far, and to the best of our knowledge, none comparison and analysis of different XAI methods have been performed on satellite multispectral imagery.

In this study, we aim to fill this gap by evaluating quantitively and qualitatively different aspects of XAI methods towards understanding black-box model decisions in multi-label RS datasets. In particular, we assessed several widely-used XAI methods, i.e., *Saliency, Input × Gradient, Integrated Gradients, Guided Backpropagation, Grad-CAM, Guided Grad-CAM, Lime, Occlusion, DeepLift* as well as their alternatives which integrate the *SmoothGrad* approach. To evaluate the studied XAI methods quantitatively, we utilized the metrics of *Max-Sensitivity, Area Under the Most Relevant First* perturbation curve, *File Size* and *Computational Time*. Moreover, we qualitatively evaluated them by assessing correctly and wrongly classified cases. We further examined and discussed the differences in the explanations for multiple labels that co-exist in the same image. To perform our experiments, we relied on the well-established Densely Connected Convolutional Network (DenseNet) (Huang et al., 2017) and Residual Neural Network (ResNet) (He et al., 2016) and trained them in challenging RS benchmark datasets, i.e., BigEarthNet (Sumbul et al., 2019) and SEN12MS (Schmitt et al., 2019), achieving state-of-the-art results. Last but not least, we offer a discussion on the gained insights regarding black-box model decisions for various prediction cases as well as datasets' composition.

## 2. Materials and Methods

In this section, we describe the employed XAI methods as well as the utilized evaluation metrics in order to quantitatively assess and compare their performance. Moreover, the two benchmark remote sensing datasets and the deep learning models are also demonstrated while their performance is evaluated. Specific implementation details are also included in the last part of this section.

*2.1. Explainable AI Methods*

In our study, we selected a broad set of the most widely used XAI methods, which are appropriate to explain individual model predictions. Also, these methods utilize various different algorithmic approaches. For the rest of the paper, let a model $f \in \mathcal{F}$ be a function $f: \mathbb{R}^D \to \mathbb{R}^C$ with input $x = (x_1, \ldots, x_D) \in \mathbb{R}^D$ and output $f(x) = (f_1, \ldots, f_C)(x) \in \mathbb{R}^C$ which is the amount of evidence for predicting a number of $C$ classes in a multiclass problem. An explanation method $\Phi: \mathcal{F} \times \mathbb{R}^D \to \mathbb{R}^D$ attributes relevance, contribution, or importance scores of the prediction $f_c(x)$ of a class to each feature $x_d$.

The *Saliency (Sal)* method (Simonyan et al., 2014) is estimated by the gradient of the output $f_c(x)$ with respect to (w.r.t.) the input $x$:

$$\Phi_{Sal}(f_c, x) = \nabla f_c(x) \tag{1}$$

Briefly, the *Saliency* approach indicates which features need to be changed the least to affect the score of a particular class the most.

The *Input × Gradient (InputXGrad)* method (Shrikumar et al., 2017) is an extension of the *Saliency* method, which uses the element-wise product of the input and the gradient:

$$\Phi_{InputXGrad}(f_c, x) = x \odot \nabla f_c(x) \tag{2}$$

Intuitively, the *Input × Gradient* product is the total contribution of each feature to the approximately linearized model's output. The implementation of both *Saliency* and *Input × Gradient* methods is straightforward; however, they are only applicable to differentiable models.

The ***Integrated Gradients (IntGrad)*** method (Sundararajan et al., 2017) is defined as the integral of the gradients along the straight-line path from a baseline $x' = (x'_1, \ldots, x'_D)$ to the input $x = (x_1, \ldots, x_D)$:

$$\Phi^d_{IG}(f_c, x) = (x_d - x'_d) \times \int_0^1 \frac{\partial f_c(\tilde{x})}{\partial \widetilde{x_d}}\bigg|_{\tilde{x} = x' + a(x - x')} da \quad \forall\, d \in \{1, \ldots, D\} \tag{3}$$

The baseline $x'$ can be an input that represents the non-appearance of a feature in the original $x$. For instance, $x'$ can be the black/ all zero image or even random noise. *Integrated Gradients* has a strong theoretical justification (Sundararajan et al., 2017) as well as it is quite easily implemented. On the other hand, this method requires a baseline reference, which is an extra input. Additionally, it can be time-consuming due to the number of samples to approximate the integral.

The ***Guided Backpropagation (GuidedBackprop)*** method (Springenberg et al., 2015) combines both *Saliency* and *Deconvnet* (Zeiler and Fergus, 2014) methods. *Guided Backpropagation* computes the gradient of the output $f_c(x)$ w.r.t. the input, except that when propagating through ReLU functions, only non-negative gradients are backpropagated. *Guided Backpropagation* implementation is not as straightforward as *Saliency*, and is applicable to differentiable models that contain ReLU functions.

The ***Grad-CAM*** method (Selvaraju et al., 2017) computes the gradients of the output $f_c(x)$ w.r.t. feature map activations $A^k$ of a given layer. At that point, the gradients are averaged for each channel $k$ (along width $W$ and height $H$ dimensions) to obtain the importance weights:

$$a_k^c = \frac{1}{H \cdot W} \sum_i^W \sum_j^H \frac{\partial f_c}{\partial A^k_{ij}}(x) \tag{4}$$

Consequently, the average gradient for each channel $a_k$ are multiplied by the layer activations $A^k$ and the results are summed over all channels. Finally, a ReLU is applied to the output, returning only non-negative attributions and bilinear interpolation can be used in order to upsample the *Grad-CAM* output to the input image resolution:

$$\Phi_{Grad-CAM}(f_c, x) = UPSAMPLE_{Bilinear}(ReLU(\sum_k a_k^c A^k)) \tag{5}$$

Intuitively, *Grad-CAM* highlights coarse regions of the image that have a positive contribution to $f_c(x)$.

The ***Guided Grad-CAM*** method (Selvaraju et al., 2017) is the element-wise product of *Guided Backpropagation* with the upsampled *Grad-CAM* attributions:

$$\Phi_{Guided\ Grad-CAM}(f_c, x) = \Phi_{Grad-CAM}(f_c, x) \odot \Phi_{GuidedBackprop}(f_c, x) \tag{6}$$

This method combines the fine-grained details of *Guided Backpropagation* with the course localization advantages of *Grad-CAM*. Both *Grad-CAM* and *Guided Backpropagation* are mainly designed for deep learning CNN models.

The ***Occlusion*** (Zeiler and Fergus, 2014) systematically replaces different contiguous rectangular patches $p_i \in P$ of the input image with a given baseline (e.g., all-zero patch) and monitors the decrease of the prediction function $f_c(x)$:

$$\Phi^{p_i}_{Occlusion}(f_c, x) = f_c(x) - f_c(x'_{p_i}) \tag{7}$$

The implementation of this method requires the least effort and can be applied to any model as it does not require access to the internal parts of the model. However, *Occlusion* requires an additional baseline input.

The ***DeepLift*** method (Shrikumar et al., 2017) is a recursive backpropagation-based method that attributes the difference in the outputs $f_c(x) - f_c(x')$ on the differences between the input $x = (x_1, ..., x_D)$ and a baseline $x' = (x'_1, ..., x'_D)$. We present *DeepLift* similar to the formulation by (Ancona et al., 2018). Specifically, *DeepLift* runs forward propagation for both original $x$ and baseline $x'$ inputs. Then, it stores the difference of the weighted activation $\Delta z_{ij} = w^{(l,l+1)}_{i,j}(x^l_i - x'^l_i)$, where $x^l_i$ and $x'^l_i$ are the activations of the neuron $i$ in the $l$ layer and $w^{(l,l+1)}_{i,j}$ is the weight between neuron $i$ and neuron j in the next $l+1$ layer. Finally, the backpropagation of the difference in the outputs to the final input contributions $\Phi^d_{DeepLift}(f_c, x)$ is achieved with the following rules:

$$r^{(L)}_i = \begin{cases} f_c(x) - f_c(x'), & i = c \\ 0, & i \neq c \end{cases} \tag{8}$$

$$r^{(l)}_i = \sum_j \frac{\Delta z_{ij}}{\sum_{i'} \Delta z_{i'j}} r^{(l+1)}_j \tag{9}$$

$$\Phi^d_{DeepLift}(f_c, x) = r^{(1)}_d, \tag{10}$$

where $L$ is the output layer and $r^{(1)}_d$ is the contribution of each feature on the input layer.
*DeepLift* has similar theoretical justifications to *Integrated Gradients* (Shrikumar et al., 2017) as well as is computationally more efficient. However, it is applicable only to deep learning models and requires an additional baseline input.

The ***Lime*** (Ribeiro et al., 2016) explanations can be achieved by approximating locally the predictions $f_c(x)$ around a specified input $x$ using a simpler self-explanatory surrogate model g (e.g., linear regression, $g(z) = w \cdot z$). *Lime*'s core idea is that the surrogate model is trained on different *interpretable f*eatures z from the original model inputs $x$. For instance, in our case, an *interpretable* representation $\tilde{z}$ is a binary vector indicating the presence or absence of a contiguous patch of the image, such that $h(\tilde{z}) = \tilde{x}$ where $h$ is a mapping between the binary vectors and the corresponding perturbed image. An interesting aspect of the *Lime* method is that each generated sample $\tilde{z}$ is weighted by the corresponding similarity of $\tilde{x}$ to the instance of interest $x$. In our case, we used an exponential kernel between the original and the perturbed input $\pi_x(\tilde{z}) = \exp(-\|x - h(\tilde{z})\|^2_F)$ as a similarity function. The surrogate model is obtained by the minimization of:

$$g = argmin_{g'} \sum_{\tilde{z}} \pi_x(\tilde{z}) \cdot ((f_c \circ h)(\tilde{z}) - g'(\tilde{z})) \tag{11}$$

The positive $\Phi^i_{LIME}(f_c, x) = w_i$ explanations (i.e., weights of the linear model $g$) highlight the regions that contribute to the prediction $f_c(x)$.
Similar to *Occlusion,* this method is model-agnostic and can be applied to any classifier or regressor. Although Lime is a modular and extensible approach, there are various parameters that have to be defined (i.e., surrogate model, image segmentation algorithm to derive the contiguous patches/ super-pixels, the similarity function $\pi_x$ and the number of generated samples).

The **SmoothGrad** (**SG**) (Smilkov et al., 2017) is a method that can be used on top of other attribution methods. *SmoothGrad* generates multiple samples by adding Gaussian noise to the original input $x$ and averages the calculated attributions:

$$\Phi_{SG}(f_c, x) = \mathbb{E}_{\varepsilon \sim \mathcal{N}(0,\sigma^2 I)}[\Phi(f_c, x + \varepsilon)] \qquad (12)$$

This approach can be understood as an averaging process that makes the initial explanation $\Phi(f_c, x)$ smoother (Samek et al., 2021). *SmoothGrad* does not require additional constraints to the underlying XAI method $\Phi$. Nevertheless, this method increases the computation time due to the number of samples.

*2.2. Evaluating Metrics*

In order to evaluate quantitatively the performance of the aforementioned XAI methods, we employed various metrics.

The **Max-Sensitivity** (**MS**) (Yeh et al., 2019) metric measures the reliability in terms of the maximum change in an explanation $\Phi(f_c, x)$ with small input perturbations $x'$ and it is estimated using Monte Carlo sampling:

$$SENS_{MAX}(\Phi, f_c, x, r) = \max_{\|x'-x\|_\infty \leq r} \|\Phi(f_c, x') - \Phi(f_c, x)\|_F, \qquad (13)$$

where $\|\cdot\|_\infty$ is the maximum norm, $\|\cdot\|_F$ is the Frobenius norm and $r$ is the input neighborhood radius. Naturally, we would not prefer an explanation to have high *Max-Sensitivity*, since that would entail differing explanations with minor variations in the input. This fact might lead us to distrust the explanations.

Another way to assess the performance of XAI methods quantitatively is the **Most Relevant First** (**MoRF**) perturbation curve (Samek et al., 2017). This procedure measures the reliability of an explanation by testing how fast the $f_c(x)$ decreases, while we progressively remove information (e.g., perturb pixels) from the input $x$ (e.g., image), that appears as the most relevant by the explanation $\Phi(f_c, x)$.

Specifically, let $x = (x_1, \ldots, x_D) \in \mathbb{R}^D$ be the input. We denote by $\Phi(f_c, x)^\downarrow \in \mathbb{R}^D$ a vector with the same components as $\Phi(f_c, x) = (\varphi_1, \ldots, \varphi_D) \in \mathbb{R}^D$, but sorted in non-increasing order. There is a permutation $\sigma: \{1, 2, \ldots, D\} \to \{1, 2, \ldots, D\}$, such that:

$$\Phi(f_c, x)^\downarrow = (\varphi_{\sigma(1)}, \ldots, \varphi_{\sigma(D)}) \in \mathbb{R}^D \qquad (14)$$

We call the permutation $\sigma$ as *argsort* and for each $j < k \Rightarrow \varphi_{\sigma(j)} \geq \varphi_{\sigma(k)}$. Also, we define the sets $S_j := \{\sigma(i) : i \leq j\}$ for $j = 1, 2, \ldots, D$. Thus, the $k$ *Most Relevant* features of $x$ are the components positioned at $S_k = \{\sigma(1), \sigma(2), \ldots, \sigma(k)\}$. Finally, if we reformulate the input as a finite sequence $x = (x_d)_{d=0}^D$, the perturbated input based on the $k$ *Most Relevant* features is:

$$x_{MoRF}^{(k)} = \left( \begin{cases} x'_d, & \text{if } d \in S_k \\ x_d, & \text{if } d \notin S_k \end{cases} \right)_{d=0}^D, \qquad (15)$$

where $x'_d$ is a perturbated feature. For instance, in the context of an image, this perturbation might be a zero, a random or an interpolated pixel value. The faster the curve $f_c(x_{MoRF}^{(k)})$ for $k = 1, 2, \ldots, D$ decreases, the more reliable the explanation is.

In order to quantify and compare the degree of this decrease, we use the **Area Under the Most Relevant First (AUC - MoRF)** perturbation curve. Based on the *trapezoidal rule*, this area is:

$$AUC_{MoRF}(\Phi, f_c, x, r) = \sum_{k=2}^{D} \frac{f_c(x_{MoRF}^{(k-1)}) + f_c(x_{MoRF}^{(k)})}{2} \quad (16)$$

Thus, we would like to minimize the $AUC - MoRF$ score for an explanation.

Moreover, the *File Size* of an XAI visualization was proposed by (Samek et al., 2021) as a quantification metric of the included amount of information. The smaller the file size, the less complex information in the results, thus potentially more concise and interpretable for a human.

The *Computation Time* is an essential metric to quantify the performance of any given algorithm. High computation time is a barrier to XAI methods' applicability as well as to their integration in complex pipelines with potential requirements for real-time performance.

*2.3. Remote Sensing Datasets and Multi-label Deep Learning Models*

Remote sensing data contain semantically complex content so that several land-cover classes co-exist simultaneously. Thus, in this study, we investigated the multi-label setup in order to assess the selected XAI methods. We employed two remote sensing benchmark datasets, i.e., BigEarthNet (Sumbul et al., 2019) and SEN12MS (Schmitt et al., 2019), which have recently gained significant attention from the research community towards developing and assessing deep learning architectures for remote sensing image understanding tasks.

In particular, BigEarthNet consists of 590,326 non-overlapping Sentinel-2 image patches, which were constructed by 125 different tiles with less than 1% cloud cover. BigEarthNet is annotated with labels provided by the CORINE Land Cover (CLC) map of 2018. However, in this paper, the alternative nomenclature with 19 classes proposed by (Sumbul et al., 2020) for BigEarthNet was used, as it better expresses single-date Sentinel-2 images.

SEN12MS consists of 180,662 triples of Sentinel-1 synthetic aperture radar (SAR) data, Sentinel-2 full multispectral imagery and MODIS-derived land cover image patches. The dataset was constructed by 252 different scenes, which are globally distributed. SEN12MS was developed based on a sophisticated workflow to avoid cloud-affected images. Each image patch consists of $256 \times 256$ pixels in size, and there is a 50% overlap between adjacent patches. In this paper, we used the SEN12MS modification for multi-label classification task. Also, we used the simplified International Geosphere–Biosphere Programme (IGBP) classification scheme, following the IEEE GRSS Data Fusion Contest 2020 (Robinson et al., 2021). This simplified scheme consists of 10 aggregated classes compared to the original IGBP with 17 classes.

For our experiments, we adopted DenseNet-121 architecture (Huang et al., 2017), in which each convolutional layer receives feature maps from all previous layers and transmits its feature maps to all subsequent layers. We modified the first layer of the DenseNet-121 to adapt to the 12 input bands (B10 Cirrus band is excluded) for both datasets, as well as the final classification layer was changed to output 19 and 10 classes for BigEarthNet and SEN12MS, respectively. During training, we employed a scheduler to reduce the learning rate when the validation set's loss has stopped decreasing. Moreover, we utilized early stopping based on the loss of the validation set. The batch size for the SEN12MS was 32 samples and for the BigEarthNet was 64. Additionally, we employed random rotations of the input images by -90°, 0°, 90°, or 180° and horizontal flips in order to augment the datasets.

For the SEN12MS dataset, which is a recent one, there was no available multi-label classification model in the literature to compare our results. For this reason, we also utilized the well-established ResNet-50 (He et al., 2016) model. To fairly compare the two models, we utilized for ResNet the same training setup as described above.

In order to evaluate the employed models, we assessed different overall and per class metrics, including $F_1$-score, Recall, Precision, Hamming Loss, Rank Loss, Coverage Error and compared our results with the literature. For BigEarthNet, the DenseNet-based model resulted in state-of-the-art performance across all overall metrics compared with recently reported efforts (Sumbul et al., 2020) (Table 1 and Table S1). For the SEN12MS dataset, the DenseNet-121 and ResNet-50 models achieved similar results to each other (Table 2 and Table S2) with overall $F_1$ scores of 74.35 % and 74.62 %, respectively.

Table 1. Comparing the performance of DenseNet121 model against the state-of-the-art in BigEarthNet (Sumbul et al., 2020). With bold the higher score per evaluation metric.

| Metric | K-Branch CNN | VGG16 | VGG19 | ResNet50 | ResNet101 | ResNet152 | DenseNet121 |
|---|---|---|---|---|---|---|---|
| $F_1$-score (%) | 72.73 | 76.01 | 75.96 | 77.11 | 76.49 | 76.53 | **82.22** |
| Recall (%) | 78.96 | 75.85 | 76.71 | 77.44 | 77.45 | 76.24 | **84.70** |
| Precision (%) | 71.61 | 81.05 | 79.87 | 81.39 | 80.18 | 81.72 | **83.60** |
| Hamming Loss | 0.093 | 0.077 | 0.079 | 0.075 | 0.077 | 0.075 | **0.061** |
| One Error | 0.103 | 0.073 | 0.071 | 0.072 | 0.082 | 0.072 | **0.029** |
| Rank Loss | 0.056 | 0.048 | 0.048 | 0.047 | 0.049 | 0.046 | **0.026** |
| Coverage Error | 4.730 | 4.603 | 4.606 | 4.613 | 4.628 | 4.552 | **3.893** |

Table 2. Comparing the performance of DenseNet121 & ResNet50 models in the SEN12MS Dataset. With bold the higher score per evaluation metric.

| | $F_1$-score (%) | Recall (%) | Precision (%) | Hamming Loss | One Error | Rank Loss | Coverage Error |
|---|---|---|---|---|---|---|---|
| **DenseNet121** | 74.35 | 82.46 | **73.80** | **0.122** | 0.138 | 0.057 | **2.928** |
| **ResNet50** | **74.62** | **83.85** | 73.66 | 0.123 | **0.118** | **0.056** | 2.943 |

*2.4. Implementation Details and Parameter Selection*

In this subsection, implementation details are discussed and presented. Towards a fair comparison, we employed the proposed parameters that the methods were published with, we tuned when required the parameters to ensure optimal performance from the studied methodologies and used the same parameters across all experiments. In particular, for the XAI methods, various parameters were tested in order to assess their overall sensitivity and performance. Based on a trial-and-error investigation, the configuration that provided the most stable and robust results was selected, in order also to accommodate in the available hardware the different size of the input images. More specifically, for the visualization of *Saliency*, *Input × Gradient* and *Integrated Gradients*, the absolute values of attributions were utilized in all our experiments. In the case of *Integrated Gradients*, we employed the black/ all zero image as a baseline reference input, following the original proposed approach. Moreover, to approximate the integral, we used 50 samples along the path. Regarding *Grad-CAM*, we used the feature maps of the last convolutional layer for computing the attributes. Furthermore, for the baseline $x'$ requirement of *DeepLift*, we used a blurred version of the initial image as was originally proposed. Considering *Occlusion*, we applied a 15 × 15 sliding window with 5 × 5 strides on BigEarthNet and a 32 × 32 window with 6 × 6 strides on SEN12MS. Also, we used the black/ all zero patches for the replacements of the baseline $x'_{p_i}$ inputs. For *Lime*, we used super-pixels that have been derived from the SLIC image segmentation algorithm (Achanta et al., 2012). Additionally, 64 segments were exploited for BigEarthNet, as well as 100 segments for SEN12MS. For each case, 6000 samples were generated in order to train the linear surrogate models. We selected to present results for the super-pixels with the highest positive weights of the linear model. When *SmoothGrad* was integrated on top of other methods, 30 samples were generated to obtain the results. Regarding the presented visualization of XAI methods, we employed colored gray-scale images that indicate the sum of attributions along the channels/ bands. Last but not least, we ran our experiments on a single RTX 3070 (8GB memory) GPU, Ryzen 7 3700X CPU, 32GB RAM and employed implementations from the Captum library (Kokhlikyan, 2020).

## 3. Experimental Results and Evaluation

The following two subsections 3.1 and 3.2 present the quantitative and qualitative evaluation of the studied XAI methods. For these experiments, we relied on DenseNet models trained on BigEarthNet and SEN12MS.

*3.1. Quantitative Evaluation*

In order to quantify the reliability of the studied XAI methods we utilized *Max-Sensitivity* and *AUC-MoRF* metrics. In particular, the assessment of XAI performance in terms of methods' sensitivity against slightly different noisy input was achieved using *Max-Sensitivity* metric. *AUC-MoRF* was also performed to examine how the progressive removal of those pixels that contributed the most to model's decision according to XAI, affected model prediction.

Additionally, the produced *File Size* in Kilobytes (KB) after JPEG compression for every XAI output was assessed in order to quantify the interpretability regarding the amount of information included in the extracted explanations. The *Computation Time*, as an aspect of methods' applicability, is also evaluated and recorded in seconds (hardware, software configuration details at Sect. 2.4). Table 3 demonstrates all aforementioned quantitative metrics for both datasets. We mention that the lower the scores the better the performance for all metrics.

**Table 3**. Quantitative Metrics (lower scores indicate higher performance for all metrics).

| Method | Max-Sensitivity BigEarthNet ($120 \times 120$) | Max-Sensitivity SEN12MS ($256 \times 256$) | AUC-MoRF BigEarthNet ($120 \times 120$) | AUC-MoRF SEN12MS ($256 \times 256$) | File Size (KB) BigEarthNet ($120 \times 120$) | File Size (KB) SEN12MS ($256 \times 256$) | Computation Time (Sec) BigEarthNet ($120 \times 120$) | Computation Time (Sec) SEN12MS ($256 \times 256$) |
|---|---|---|---|---|---|---|---|---|
| **Sal** | 0.55 | 0.27 | 21.47 | 36.92 | 6.22 | 23.35 | 0.07 | 0.08 |
| **Sal w. SG** | 0.24 | 0.14 | 19.57 | 36.32 | 5.71 | 22.54 | 1.75 | 1.94 |
| **InputXGrad** | 0.59 | 0.30 | 26.15 | 38.80 | 5.70 | 18.07 | 0.05 | 0.06 |
| **InputXGrad w. SG** | 0.29 | 0.16 | 25.90 | 38.67 | 5.43 | 17.68 | 0.14 | 0.24 |
| **IntGrad** | 0.38 | 0.26 | 25.64 | 37.40 | 5.07 | 17.91 | 0.23 | 0.36 |
| **IntGrad w. SG** | 0.19 | 0.13 | 25.50 | 37.64 | 4.86 | 16.72 | 4.47 | 7.96 |
| **Guided Backprop** | 0.36 | 0.23 | 23.14 | 34.67 | 5.50 | 22.68 | 0.06 | 0.07 |
| **Grad-CAM** | **0.14** | **0.03** | 16.38 | **23.62** | **1.89** | 6.31 | **0.03** | **0.03** |
| **Guided Grad-CAM** | 0.50 | 0.23 | 23.66 | 35.71 | 4.57 | 16.55 | 0.08 | 0.10 |
| **DeepLift** | 0.42 | 0.25 | 23.41 | 34.65 | 5.65 | 19.33 | 0.11 | 0.13 |
| **Occlusion** | **0.14** | **0.03** | 16.32 | 28.87 | 3.02 | **3.84** | 8.73 | 29.71 |
| **Lime** | 0.20 | 0.11 | **14.39** | 27.65 | 3.24 | 7.91 | 5.80 | 18.67 |

Regarding *Max-Sensitivity* metric, *Occlusion*, *Lime* and *Grad-CAM* achieved the lowest scores, i.e., less than 0.20 for BigEarthNet and less than 0.11 for SEN12MS. On the other hand, *Input × Gradient* presented the highest scores for both datasets (i.e., 0.59 for BigEarthNet and 0.30 for SEN12MS). Indeed, *Input × Gradient* is expected to be the most affected by slightly noisy input, since this method is highly related to the input. For all cases, the integration of *SmoothGrad* approach led to lower *Max-Sensitivity* scores than the uncustomized XAI methods (i.e., *Saliency*, *Input × Gradient* and *Integrated Gradients)*. This fact is reasonable, due to the averaging process of *SmoothGrad*.

Moreover, Figure 1 demonstrates *MoRF* perturbation curve for all XAI methods for both datasets. At each iteration, 20% of remained pixels are removed based on XAI explanations until the whole information/ pixels of the image are removed. The removed pixels are imputed based on nearest interpolation. It is expected that XAI methods that indicate more precisely those image pixels/ regions that bind model's decision will be the ones that will rapidly lose their prediction performance (mean output score) as those pixels get more and more perturbated (iterations). Random explanation (pink dashed curve) is also presented as a baseline since it is expected that by perturbating randomly image pixels, the performance will be decreased at a slower pace.

According to Figure 1a,b *Occlusion* (purple dashed curve), *Lime* (yellow dashed curve) and *Grad-CAM* (black dashed curve) achieved the fastest decrease in their curves. This observation is also quantitatively demonstrated in Table 3. In particular, *Occlusion*, *Lime* and *Grad-CAM* resulted in less than 16.38 for BigEarthNet and less than 28.87 for SEN12MS regarding the *AUC-MoRF* metric. On the contrary, *Input × Gradient* presented the nearest curve to the random baseline curve (Figure 1a,b), as well as achieved the highest *AUC-MoRF* scores (i.e., 26.15 for BigEarthNet and 38.8 for SEN12MS) (Table 3). This is potentially attributed to the fact that pixel-level methods with high *AUC-MoRF* scores such as *Input × Gradient* focus on individual pixels to which the model is the most sensitive, without managing to identify entirely comprehensively the relevant patterns and regions in the input images. Additionally, the integration of *SmoothGrad,* for the most cases, led to slightly lower *AUC-MoRF* scores and consequently more reliable explanations.

Concerning *File Size*, the results were positively correlated with the *Max-Sensitivity* results (r=0.69, p-value < 0.05 for BigEarthNet and r=0.82, p-value < 0.005 for SEN12MS). Specifically, *Occlusion, Lime* and *Grad-CAM* presented the smallest *File Size* (i.e., < 3.24 KB for BigEarthNet and < 7.91 KB for SEN12MS), as they are the lowest resolution methods. Therefore, these methods can provide rough and concise localization information to the user. On the other hand, all other methods provide higher resolution and more detailed explanations than *Occlusion, Lime* and *Grad-CAM,* leading to larger file sizes. Overall, there is a trade-off between fine-grained details and the vast amount of potentially confusing information to the user.

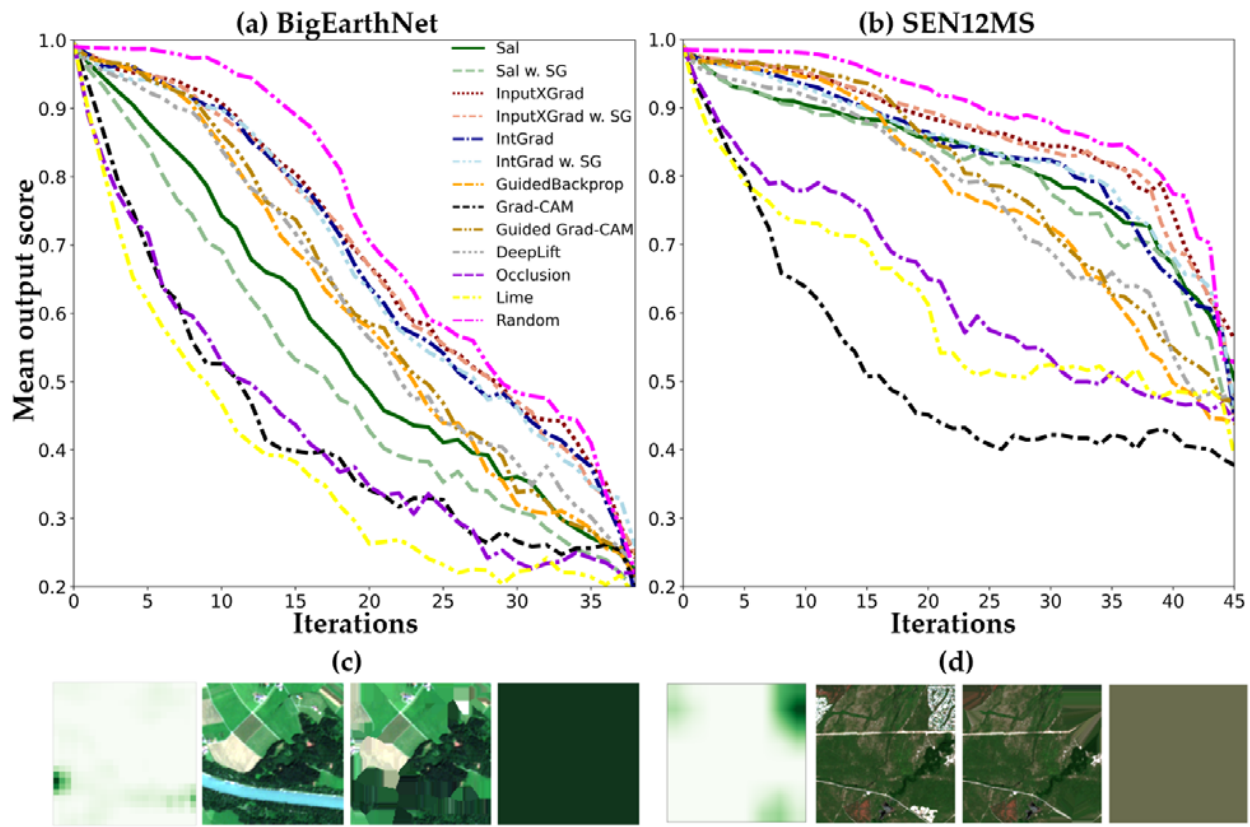

**Figure 1.** *MoRF* perturbation curves for all XAI methods for (a) BigEarthNet and (b) SEN12MS datasets. Curves indicate the drop in model's performance (mean output sigmoid probability scores) for each iteration that an additional 20% of image pixels are perturbated based on XAI explanations. (c) Example for *Inland Waters* class prediction; from left to right: *Occlusion* explanations, S2 RGB image, intermediate iteration, final iteration. (d) Example for *Urban/Built-up* class prediction; from left to right: *Grad-CAM* explanations, S2 RGB image, intermediate iteration, final iteration.

According to Table 3, except for *Saliency w. SG, Integrated Gradients w. SG, Occlusion* and *Lime,* all the other methods required less than 0.36 sec for a single execution. The selection of parameters in the aforementioned methods affects their efficiency, and consequently the *Computation Time*. More specifically, in the case of *Integrated Gradients* with *SmoothGrad,* the bottleneck was the large number of samples that proceeded in multiple GPU batches (i.e., 4.47 sec for BigEarthNet and 7.96 sec for SEN12MS). Moreover, the *Computation Time* of *Lime* and *Occlusion* was affected by the number of the generated samples and occluded images, as well as by the required time to compute $f_c(x')$ for each sample. By using a small stride length towards high-resolution explanations in the case of *Occlusion* method, *Computation Time* further increased (i.e., 8.73 sec for BigEarthNet and 29.71 sec for SEN12MS).

Additionally, only *Occlusion* and *Lime* presented a significant increase in the required computation time from the 120 × 120 pixels (BigEarthNet) to the 256 × 256 pixels (SEN12MS). This fact indicates a lower capacity against scalability. Regarding *Occlusion,* this increase is mainly due to the additional generated occluded samples based on different stride size selection. On the other hand, for *Lime,* we generated 6000 samples for both datasets; hence the only overhead was the additional required time for the evaluation of $f_c(x')$ for each sample $x'$.

It is worth mentioning that although the lowest *Max-Sensitivity* scores as well as the smallest *File Sizes* were derived from the *Occlusion* and *Grad-CAM* methods, only *Grad-CAM* required low *Computation Time* regardless of the image size.

Finally, in order to examine if the evaluation of the studied XAI methods is independent of the adopted model, we further experimented with the trained ResNet-50 on SEN12MS dataset. The results are presented in Table S3 in the supplementary material, revealing that XAI assessment is quite similar for both ResNet and DenseNet models. In particular, regarding the *Max-sensitivity* score, *Grad-CAM* and *Occlusion* still have the best performance; however, the scores in the case of ResNet (0.05 and 0.06) are slightly higher than DenseNet. Regarding the *AUC-MoRF*, *Grad-CAM* achieved once more the lowest score (25.70). Concerning the *File Size*, again the methods with the smallest file size were *Grad-CAM, Occlusion* and *Lime*.

*3.2 Qualitative Evaluation*

*3.2.1. Explaining Single Class Correct Predictions*

Initially, we studied numerous cases in both datasets that the DenseNet models managed to predict correctly the underlying classes based on the ground truth. It should be noted that we also performed our own verification process, which was based on an intensive image interpretation supported by very high-resolution satellite images derived from Microsoft Bing. Specifically, two experts initially inspected the Sentinel-2 data, the associated labels as well as image interpretation features like the size, shape, texture and the spectral signatures of the depicted objects. To further clarify the ambiguous cases, we examined the corresponding very high-resolution satellite images (Microsoft Bing). We have to mention that for all figures that demonstrate the XAI outputs, *Max-Sensitivity* score is presented in parenthesis after the corresponding method's name.

Two indicative cases that the models managed to accurately predict the class are presented in Figures 2, 3 and corresponding Figures S1 and S2 in the supplementary material. In Figure 2 (and S1), the resulted explanations for predicting the class *Urban Fabric* are presented for the BigEarthNet dataset. The model managed to accurately predict *Urban Fabric* with a 0.99 sigmoid probability score.

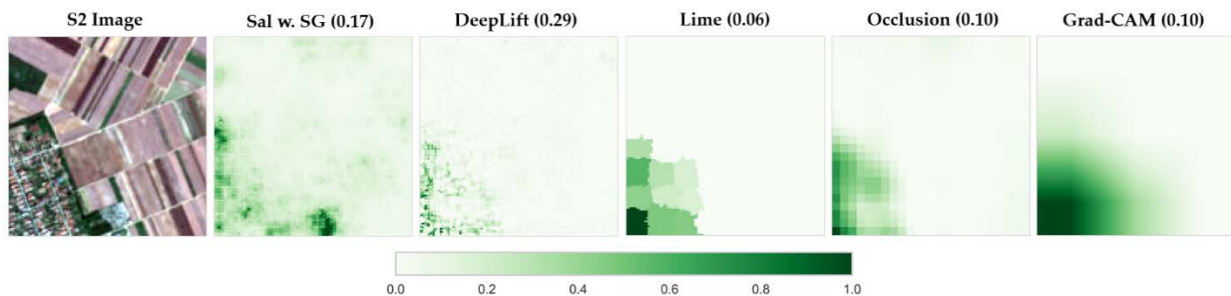

**Figure 2.** Explaining the Predictions of DenseNet for the class of *Urban Fabric* in the BigEarthNet dataset (Image ID: S2A_MSIL2A_20171002T094031_53_67).

In particular, after a close look, one can observe that all XAI methods managed to explain and locate the image-regions (i.e., South-West) that correspond to the *Urban Fabric* class. Overall, the studied methods *Lime*, *Occlusion*, and *Grad-CAM* were the most insensitive w.r.t. the input with *Max-Sensitivity* score less than 0.10. This fact indicates that these methods are reliable for the specific case. On the opposite side, *Saliency* and *Input × Gradient* were the most sensitive ones (with a *Max-Sensitivity* score of more than 0.65). Moreover, when the *SmoothGrad* approach was integrated, all methods resulted in lower *Max-Sensitivity* scores indicating that *SmoothGrad* contributes to a more robust and insensitive outcome. However, after a close look and visual examination, no significant differences were observed, apart from the fact that a certain amount of noise was eliminated. For the *Integrated Gradients* method, adding *SG* did not result in any difference since the method already delivers smooth results due to its inherent computations (Figure 2, Figure S1).

In a similar manner, in Figure 3 (and corresponding Figure S2), the extracted explanations for the class *Water* are presented for the SEN12MS dataset. The model accurately predicted *Water* with a 0.99 sigmoid probability score. In particular, experimental results indicate that *Occlusion*, *Lime*, *Grad-CAM*, *Saliency*, *Input × Gradient*, *Integrated Gradients* and the corresponding methods with *SmoothGrad* presented interpretable explanations, as they successfully identified the water region in the image (i.e., the West and North-West region). Instead, *DeepLift*, *Guided Backpropagation*, *Guided Grad-CAM* were not able to interpret DenseNet's decision adequately. More specifically, *Guided* methods focused on irrelevant regions of the image that contained striking image features like edges (i.e., urban area). Regarding *Max-Sensitivity*, the lowest scores were achieved by *Occlusion* (0.01), *Grad-CAM* (0.01), and *Lime* (0.08) methods. Additionally, Figure S2 shows that all methods resulted in lower *Max-Sensitivity* scores and visually smoother explanations when the *SmoothGrad* approach was integrated.

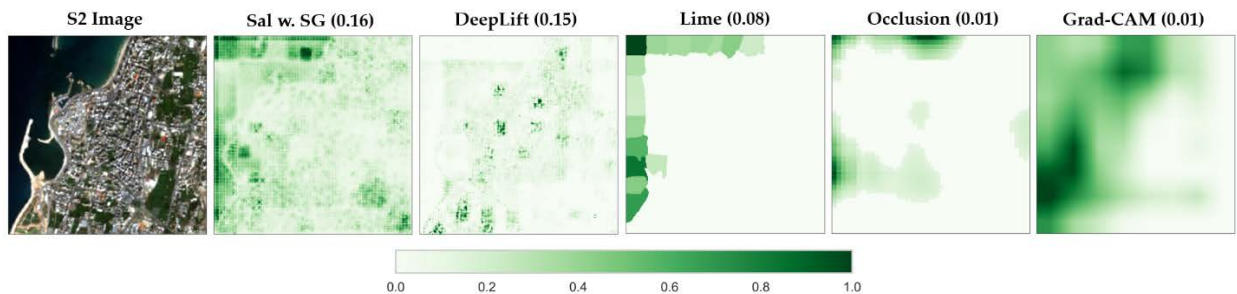

**Figure 3.** Explaining the Predictions of DenseNet for the class of *Water* in the SEN12MS dataset (Image ID: ROIs1970_fall_s2_112_p727).

*3.2.2. Explaining Correct Predictions for Multiple Competing Classes*

Furthermore, we studied several cases that the DenseNet models successfully predicted the underlying classes for cases that multiple competing classes were labeled in the same image. Two indicative cases for BigEarthNet are presented in Figures 4, 5 and corresponding Figures S3a, S3b, S4a, S4b in the supplementary material.

More specifically, in Figure 4 (and corresponding S3a and S3b), the derived visual explanations after the classification of two semantically-diverse classes, i.e., *Urban Fabric* and *Broad-leaved Forest* are presented. The model managed to accurately predict both *Urban Fabric* and *Broad-leaved Forest* with a higher than 0.99 sigmoid probability score. Overall, we observed that *Occlusion*, *Lime* and *Grad-CAM* methods were the most interpretable and sensitive w.r.t. different classes, as they presented accurate localization information to the user. *Guided Backpropagation* failed to explain DenseNet's decision against the two labels since it focused on the same image regions for both classes, indicating that it is less reliable. *Guided Grad-CAM* was slightly more sensitive (as it utilizes *Grad-CAM*) but still underperformed. *DeepLift*, *Saliency*, *Input × Gradient*, and *Integrated Gradients* were sensitive to each different label and provided different explanations. Nevertheless, in the *Urban Fabric* class, they provided more informative results than *Broad-leaved Forest*.

Moreover, in order to evaluate the performance of the studied XAI methods, we examined cases with competing classes that belong semantically to the same family and exist concurrently in the same image. In Figure 5 (and corresponding Figures S4a, S4b), an indicative case from BigEarthNet is presented with the classes *Urban Fabric* and *Industrial Units* that belong to the same Artificial Surfaces super-class. The model accurately predicted both *Urban Fabric* and *Industrial Units* with 0.90 and 0.64 sigmoid probability scores, respectively. Overall, derived explanations revealed that the model correctly focused on the North image region for the *Urban Fabric* and on the North-West area for the *Industrial Units*. In particular, all methods managed to focus on the actual image regions that correspond to considered labels. However, *Grad-CAM* and *Integrated Gradients*, when explaining decisions related to the *Urban Fabric* class, focused additionally on pixels/sub-regions of the *Industrial Units* class. Compared to previous examples, *Guided Backpropagation* was slightly more sensitive in this case (Figure S4a, S4b).

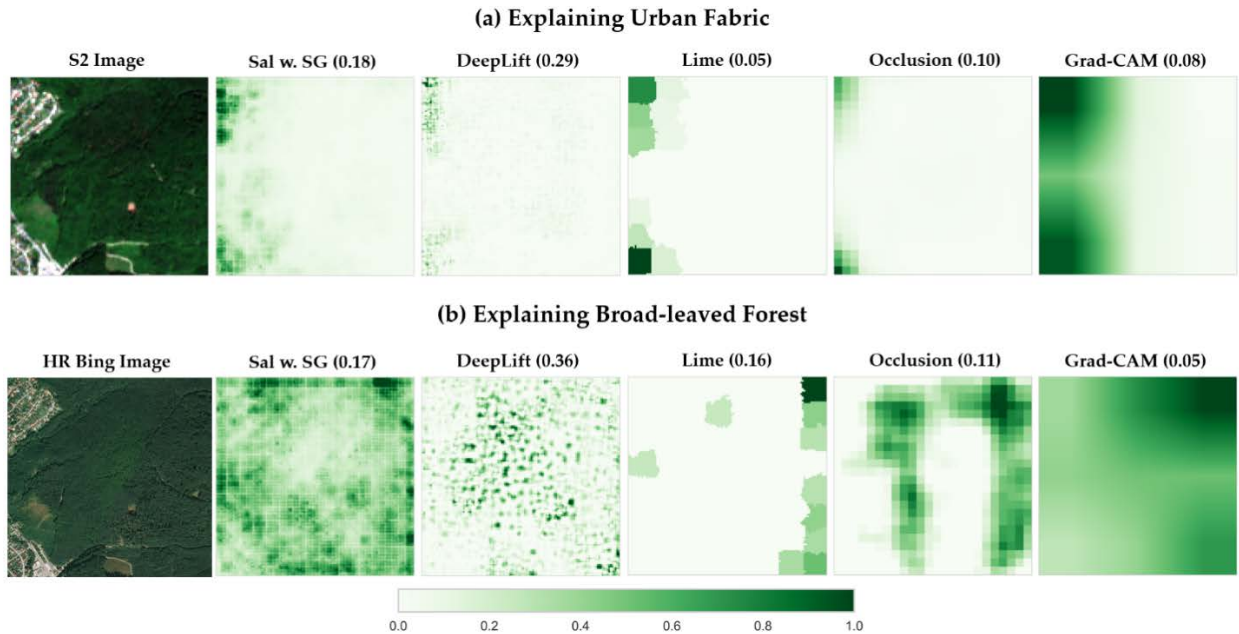

**Figure 4.** Explaining the Predictions of DenseNet for the class of (a) *Urban Fabric* and (b) *Broad-leaved Forest* in BigEarthNet dataset (Image ID: S2A_MSIL2A_20180506T100031_72_48).

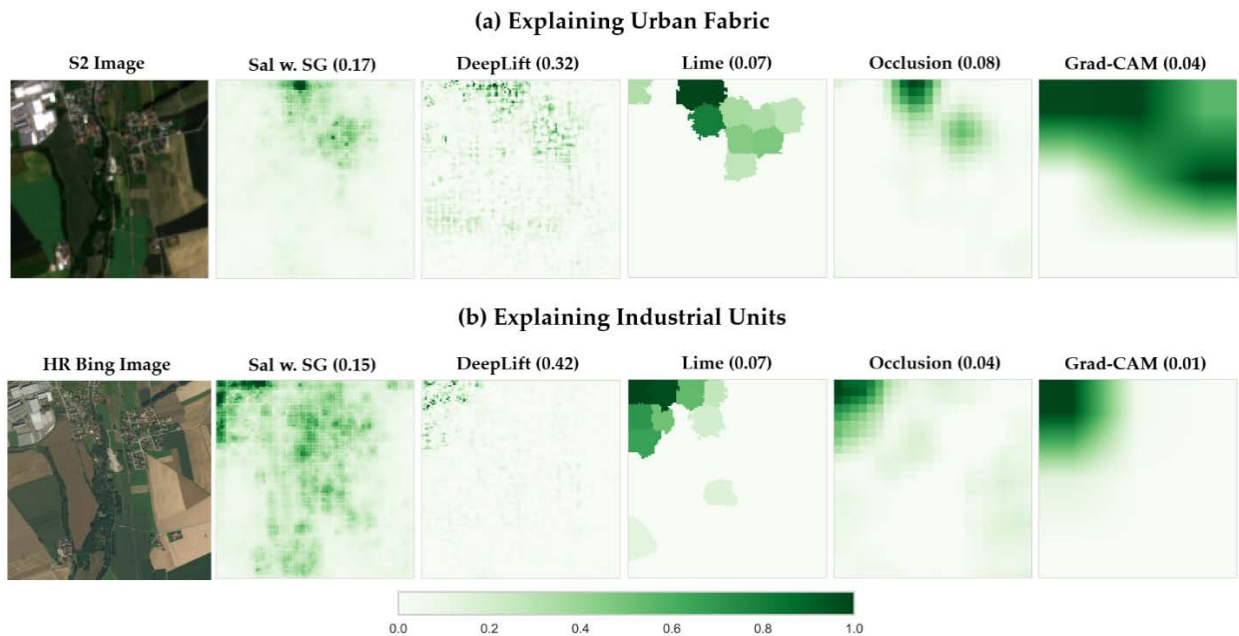

**Figure 5.** Explaining the Predictions of DenseNet for the class of (a) *Urban Fabric* and (b) *Industrial Units* in BigEarthNet dataset (Image ID: S2A_MSIL2A_20170613T101031_47_45).

Additionally, another indicative case for the explanations of *Forest* and *Savanna* class decisions on SEN12MS dataset is demonstrated in Figure 6 (and corresponding Figures S5a, S5b). The model accurately predicted both *Forest* and *Savanna* with 0.97 and 1.00 sigmoid probability scores, respectively. After visual inspection, we observed that all methods provided valuable information for *Forest* class prediction to the user. *Occlusion* and *Grad-CAM* were the most interpretable. However, only *Grad-CAM* managed to highlight the *Savanna* class region, indicating that this method is the most class-discriminative one. It is worth mentioning that *Savanna* class identification is challenging also for an expert, as *Savanna* consists of a mixed ecosystem and may span spatially the entire image.

### 3.2.3. Explaining Model Failure and Failure due to Inexact Labeling

Furthermore, we examined several cases that the models failed to predict the correct classes, which were included in the ground truth/ testing set and verified by us as well.

In particular, as Figure 7 demonstrates (as well as the corresponding Figure S6), the DenseNet model incorrectly predicted *Coniferous Forest* as a label in this particular image. Indeed, the model focused on the dark green area (i.e., South-East region), which is probably a recently irrigated crop field and misclassified it as a *Coniferous Forest*, based on the relatively darker intensity values. All XAI methods agreed with each other and focused on this particular image region.

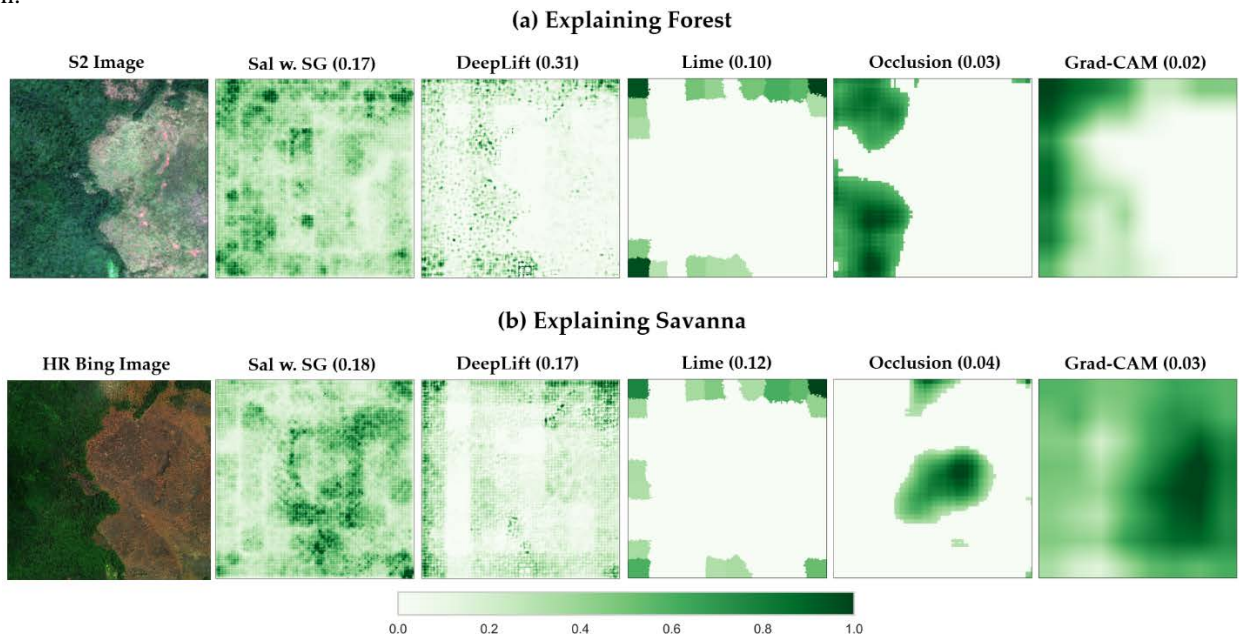

**Figure 6**. Explaining the Predictions of DenseNet for the class of (a) *Forest* and (b) *Savanna* in SEN12MS dataset (Image ID: ROIs1970_fall_s2_35_p297).

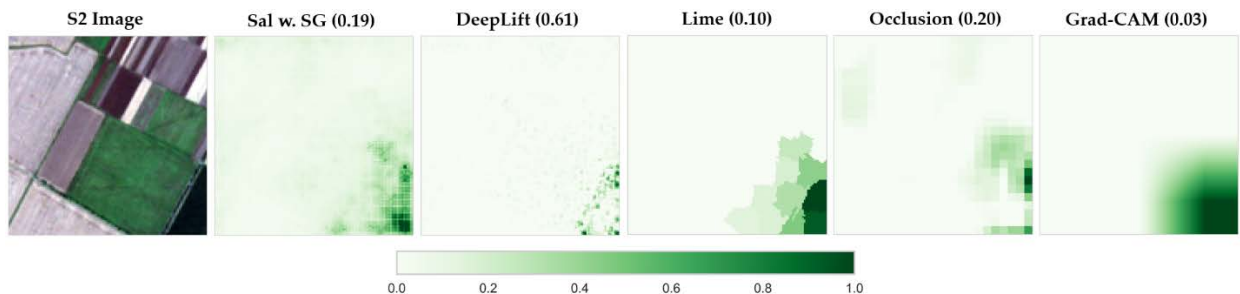

**Figure 7.** Explaining the Predictions of DenseNet for *Coniferous Forest* in BigEarthNet (Image ID: S2A_MSIL2A_20171002T094031_64_36).

Another indicative case is demonstrated in Figures 8 and S7. The model predicted the class *Water* with high confidence (0.95 sigmoid probability). Indeed, although it seems that a stream is crossing this highly dense urban area, *Water* is correctly not part of the ground truth due to its size and width. This misclassification case was due to the extended shadows that are covering the image. In particular, explanation methods indicated that the model was mainly focused on this darker cover with cloud shadows and not on any detected stream, river or water area. Except for *DeepLift* and *Guided Backpropagation*, the rest of the methods are explaining this decision quite successfully. Similar images with cloud shadows from the same region (e.g., ROIs1970_fall_s2_116) were also investigated to confirm our results. We have to mention that only a few images with cloud shadows were included in the dataset; thus, the model could not generalize adequately in cases with cloud shadows.

During the intensive examination of numerous model prediction cases and in particular, cases that model prediction and labeling were not consistent, we also focused on cases where inexact labeling in the testing datasets occurred. For instance, *Urban/ Built-up* class is not included in labeling in an image of the SEN12MS dataset (Figures 9 and S8). However, the model managed to predict *Urban/ Built-up* with a 0.97 sigmoid probability score. All XAI methods agreed with each other and correctly indicated the actual urban areas in the image (North-West and South-East regions).

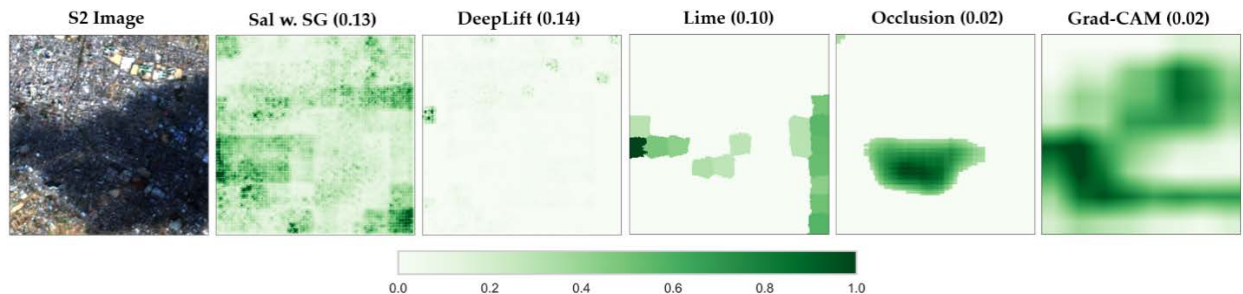

**Figure 8.** Explaining the Predictions of DenseNet for *Water* in SEN12MS (Image ID: ROIs1970_fall_s2_116_p703).

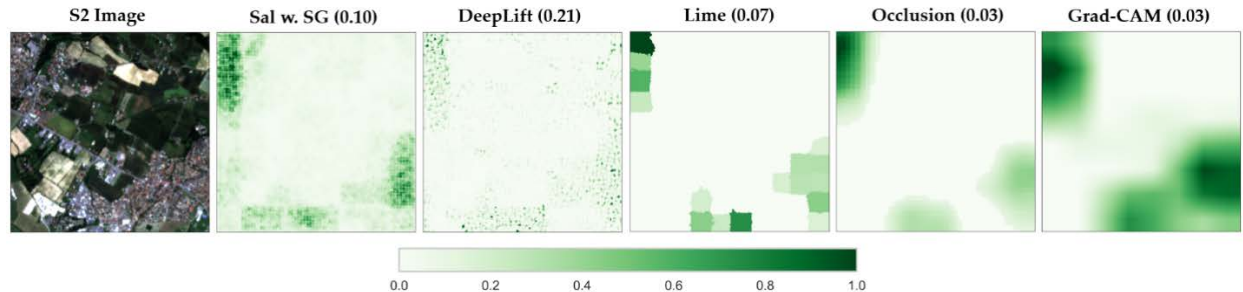

**Figure 9.** Explaining the Predictions of DenseNet for *Urban/ Built-up* in SEN12MS (Image ID: ROIs1158_spring_s2_148_p197).

## 4. Discussion

*4.1. Performance of Explainable AI methods for Multi-label Classifications Tasks*

According to our findings, *Occlusion, Grad-CAM* and *Lime* were the most interpretable XAI methods, as well as the ones that were able to explain competing multi-class decisions (class-discriminative) by also locating the corresponding image regions successfully. Quantitatively, these three methods achieved relatively low sensitivity (*Max-Sensitivity*) scores, which lead to trust their explanations (high reliability). *AUC-MoRF* metric results also confirmed that these methods are reliable. Additionally, they produced the smallest output file sizes and thus were likely more interpretable, which is in accordance with the literature (Samek et al., 2021).

On the other hand, *Guided Backpropagation* was the least reliable method as it was insensitive concerning different classes. For several cases, it focused on image regions with striking primitives (e.g., edges) independently of the predicted class, leading to outputs almost identical to the ones of an edge detector. This behavior was also observed, to a lesser extent, from the *Saliency, Input × Gradient* and *Integrated Gradients* methods. Similar observations and shortcomings have also been reported by (Adebayo et al., 2018) through their salinity checks, consisting of model parameters and data randomization tests. For classes that contain sharp edges, all methods may seem visually reliable highlighting these image features; however, this can be misleading as, e.g., *Guided Backpropagation* keeps highlighting the same image regions independently of the considered thematic classes.

Regarding qualitative evaluation, *Saliency, Input × Gradient, Integrated Gradients* and *DeepLift* in certain cases failed to provide accurate localization for classes that were spatially distributed in the image (e.g., *Savanna*). Furthermore, experiments made with *SG* indicated that methods that integrated this approach achieved lower sensitivity scores than default methods. Visually there were no significant differences in the explanations; however, noise in the default methods' outputs from irrelevant image regions was reduced. This finding is in accordance also with (Smilkov et al., 2017).

Overall, we observe that there is not a single method that stands out as the best one. *Grad-CAM* is highly reliable, interpretable, scalable, and requires less computational time but does not provide high-resolution outputs. Both *Lime* and *Occlusion* are highly reliable, interpretable and model agnostic with moderate resolution outputs but with relatively high computational time and low scalability.

*4.2. XAI for further insights in black-box models and benchmark datasets*

Through our study, the adopted models resulted in state-of-the-art performance in all evaluated metrics for both datasets. We have to mention though, that classes with the highest $F_1$ scores or high probability predictions for the examined cases did not necessarily lead to straightforward interpretable explanations (e.g., localize the corresponding image regions for a given class). For instance, in Figure S5b from SEN12MS dataset, despite the high 1.00 sigmoid probability score and the high overall $F_1$ score 84.65% for *Savanna* class (Table S2), XAI models resulted in relatively poor explanations regarding the location of the particular class in the image.

XAI methods contributed with further insights towards understanding models' predictions due to datasets particularities like training set class distribution. More specifically, by studying results and explanations for BigEarthNet, we observed that, when the model correctly predicted *Marine Water* class, occasionally, it was also focused on *Beaches, Dunes, Sands*. At the same time, for several cases, the model could not successfully predict *Beaches, Dunes, Sands* area (e.g., Figure S9). Additionally, $F_1$ score for this class was the third lowest (i.e., 63.30%) (Table S1). This fact is probably attributed to two highly correlated reasons. Firstly, a high percentage (i.e., 79%) of images labeled with *Beaches, Dunes, Sands* also include the *Marine Water* label (Figure S10). Secondly, a significant number of images that both *Marine Water* and *Beaches, Dunes, Sands* classes were depicted, were not properly labeled with *Beaches, Dunes, Sands*. These findings are also in-line with a recent study (Aune-Lundberg and Strand, 2021), which indicates that small beaches and dunes were omitted from the *Beaches, Dunes, Sands* class of CORINE Land Cover (BigEarthNet labels), while a considerable area with ocean/water was assigned in the *Beaches, Dunes, Sands* class.

## 5. Conclusions

To sum up, we evaluated XAI methods for interpreting black-box predictions derived from deep learning models with state-of-the-art performance in multi-label RS benchmark datasets. In order to quantitatively evaluate XAI performance, *Max-Sensitivity, Area Under the Most Relevant First* perturbation curve*, File Size* and *Computational Time* metrics were utilized. Extensive experiments were performed to qualitatively examine and assess the function of the considered methods. We further investigated different aspects of XAI methods regarding their applicability and explainability. Through our evaluation procedure, we found that none of the XAI methods stands out as the best one. *Occlusion, Grad-CAM* and *Lime* were the most interpretable and reliable XAI methods presenting the lowest *Max-Sensitivity* (0.14, 0.14, 0.20 for BigEarthNet, respectively) and *AUC-MoRF* scores (28.87, 23.62, 27.65 for the SEN12MS, respectively). However, none of them provides high-resolution outputs and apart from *Grad-CAM* (0.03 sec), both *Lime* and *Occlusion* are not computationally efficient (8.73 sec and 5.80 sec for BigEarthNet, respectively). Overall, our findings indicate that XAI provides valuable insights for deep black-box models' performance and decisions as well as benchmark datasets' composition and shortcomings.

**Acknowledgments**: This work has been partially supported by NEANIAS, funded by the European Union's Horizon 2020 research and innovation programme, under grant agreement No 863448. Part of this work was also funded by the Operational Program "Competitiveness, Entrepreneurship and Innovation 2014-2020" (co-funded by the European Regional Development Fund) under the project Τ6ΥΒΠ-00153 'SOSAME'.

**Conflicts of Interest**: The authors declare no conflict of interest.

# Supplementary Material

Evaluating Explainable Artificial Intelligence Methods for Multi-label Deep Learning Classification Tasks in Remote Sensing

Ioannis Kakogeorgiou* and Konstantinos Karantzalos

Remote Sensing Laboratory, National Technical University of Athens, Zographou, 15780, Greece; karank@central.ntua.gr
Athena Research Center, Athens, Greece
* Correspondence: gkakogeorgiou@central.ntua.gr; Tel.: +302107721673

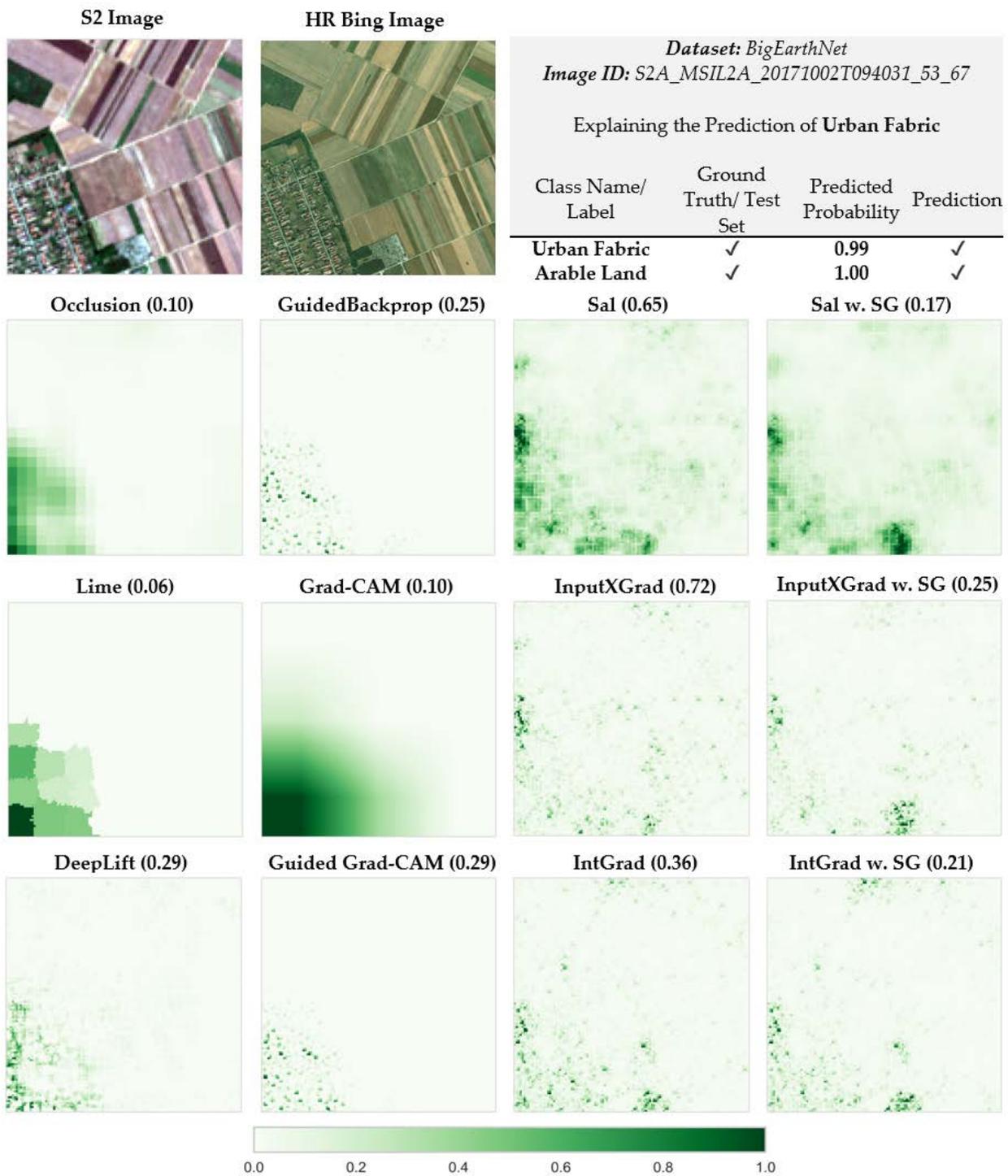

**Figure S1**. Explaining the Predictions of DenseNet for the class of *Urban Fabric* in BigEarthNet dataset (Image ID: S2A_MSIL2A_20171002T094031_53_67).

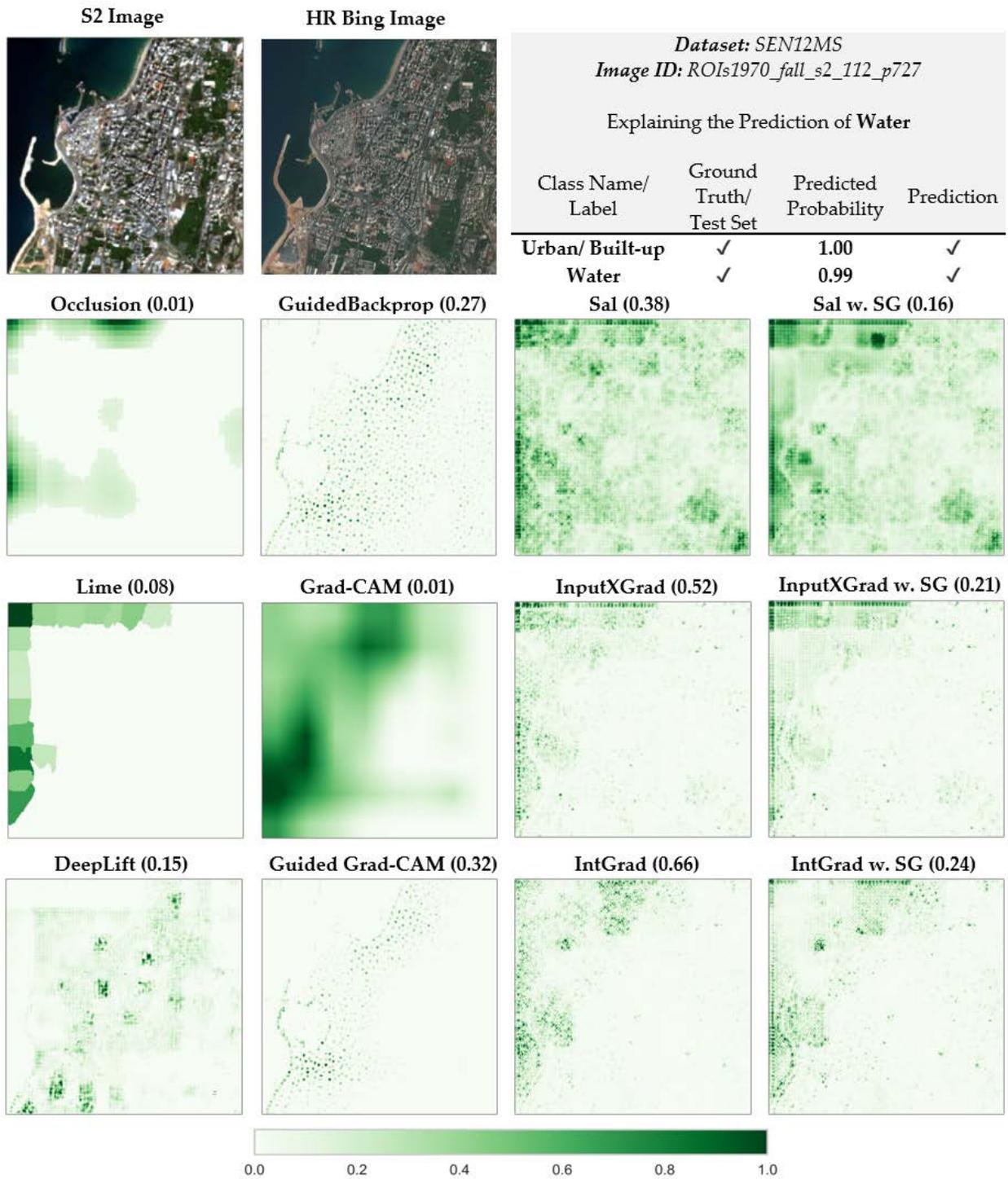

**Figure S2.** Explaining the Predictions of DenseNet for *Water* in SEN12MS (Image ID: ROIs1158_spring_s2_17_P110).

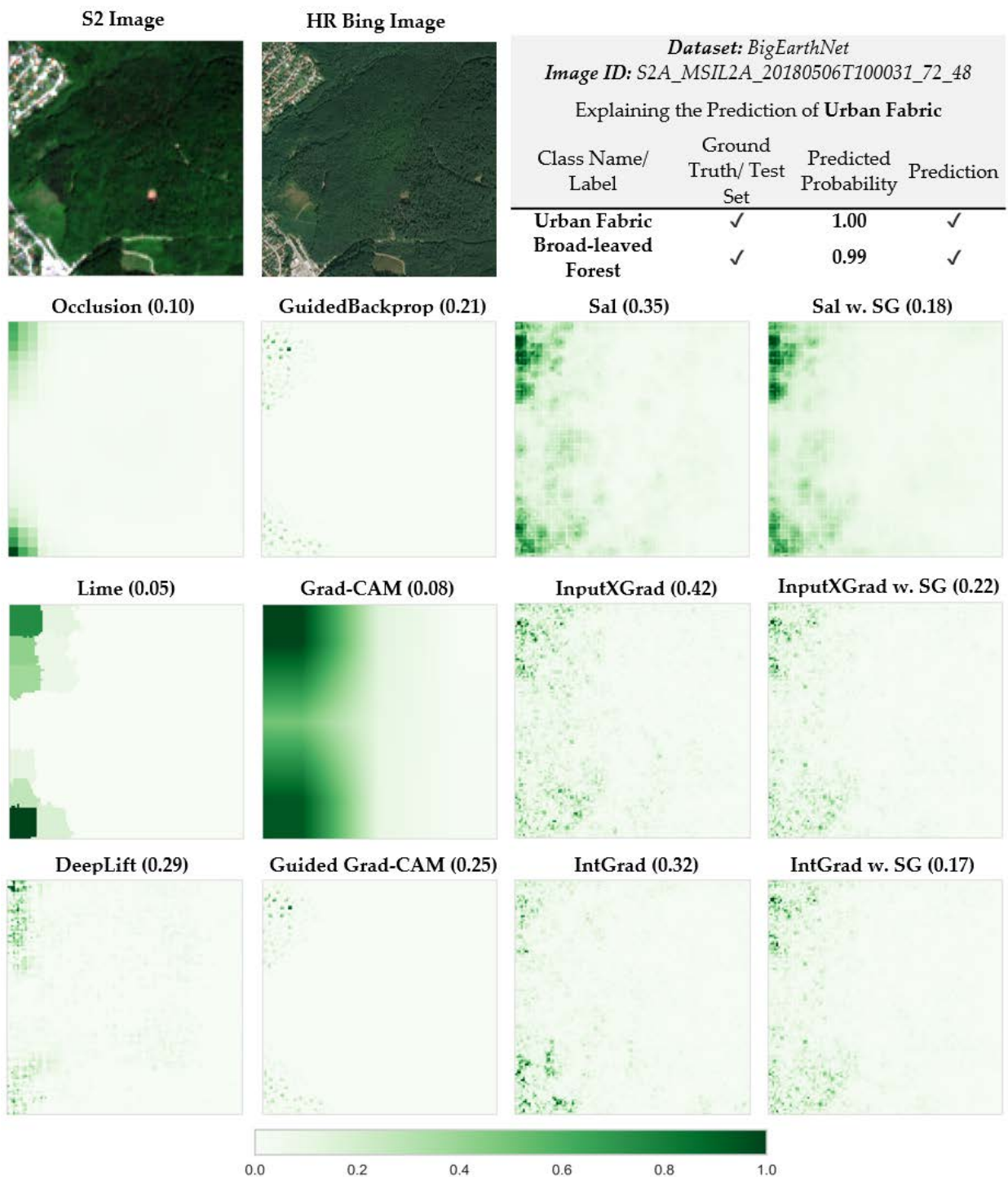

**Figure S3a.** Explaining the Predictions of DenseNet for *Urban Fabric* class in BigEarthNet dataset (Image ID: S2A_MSIL2A_20180506T100031_72_48).

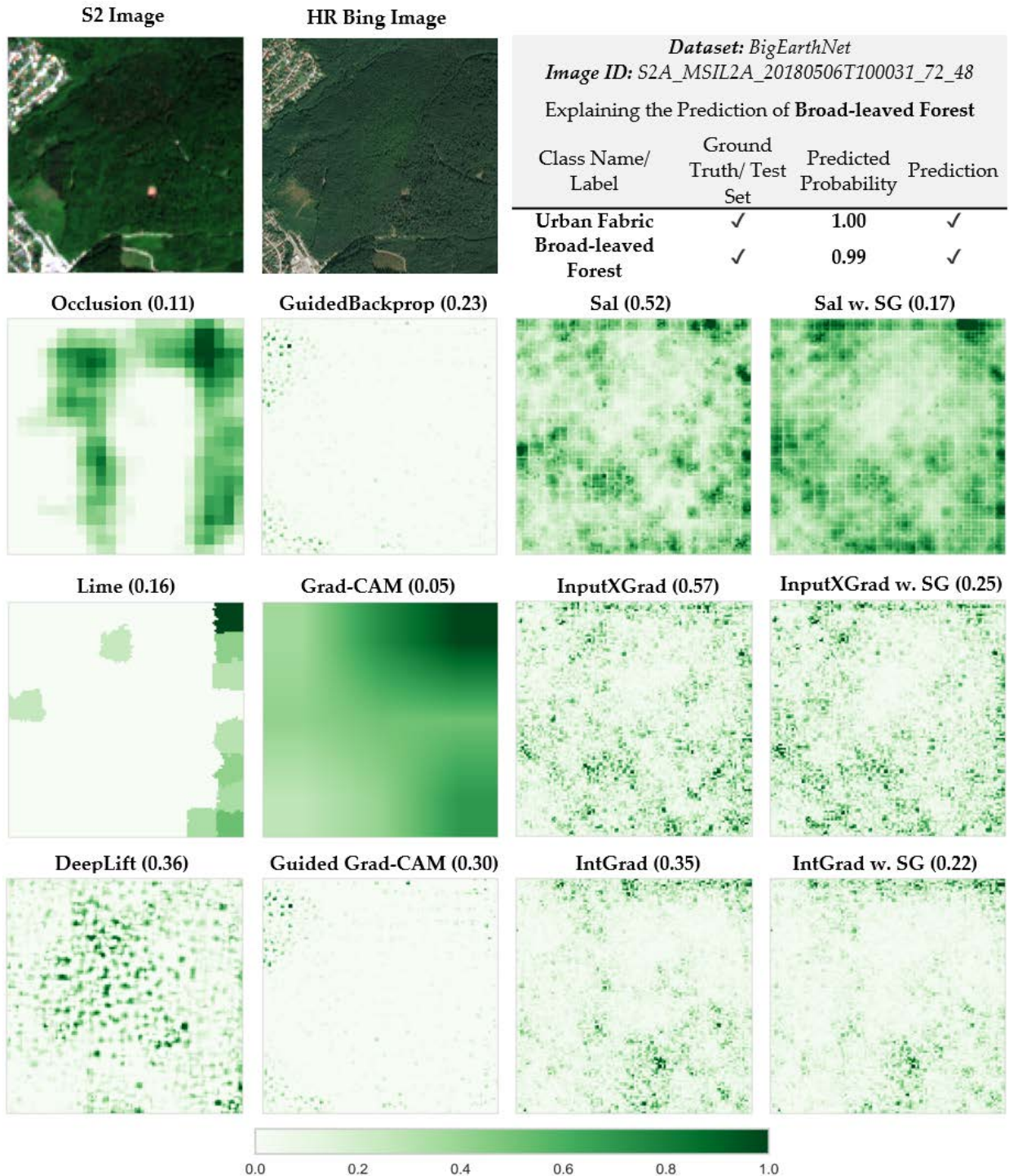

**Figure S3b.** Explaining the Predictions of DenseNet for *Broad-leaved Forest* class in BigEarthNet dataset (Image ID: S2A_MSIL2A_20180506T100031_72_48).

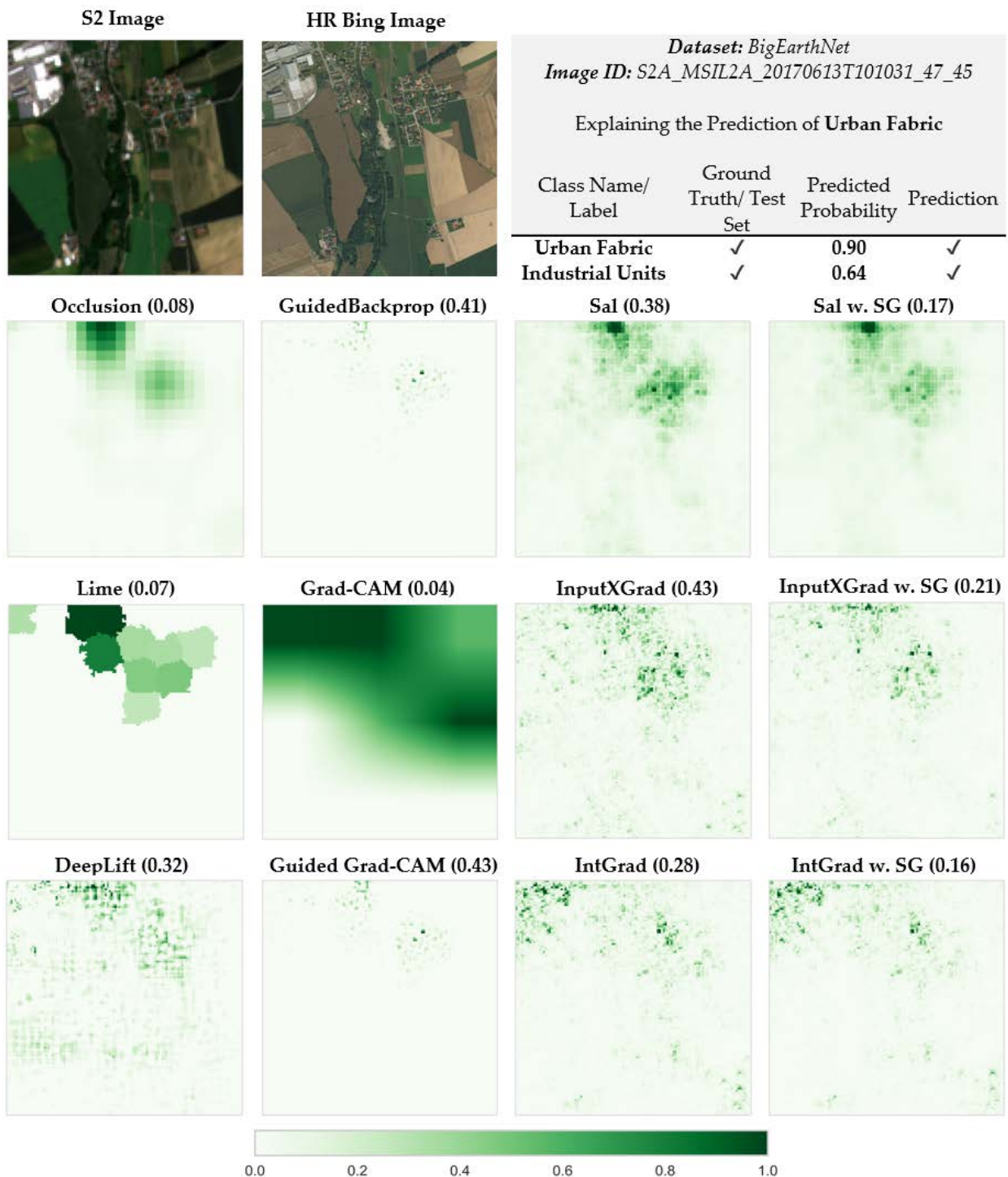

**Figure S4a.** Explaining the Predictions of DenseNet for *Urban Fabric* in BigEarthNet (Image ID: S2A_MSIL2A_20170613T101031_47_45).

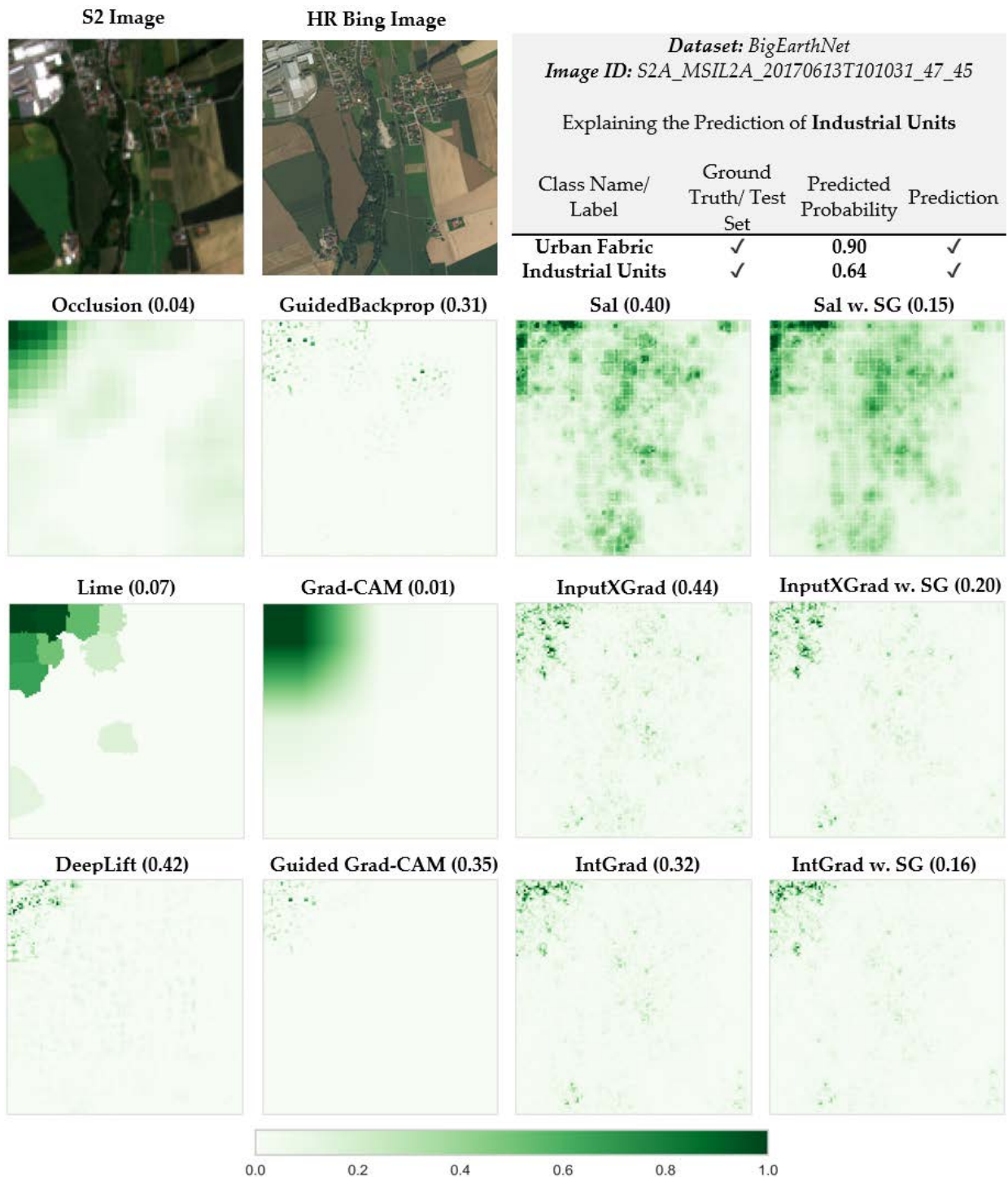

**Figure S4b.** Explaining the Predictions of DenseNet for *Industrial Units* in BigEarthNet (Image ID: S2A_MSIL2A_20170613T101031_47_45).

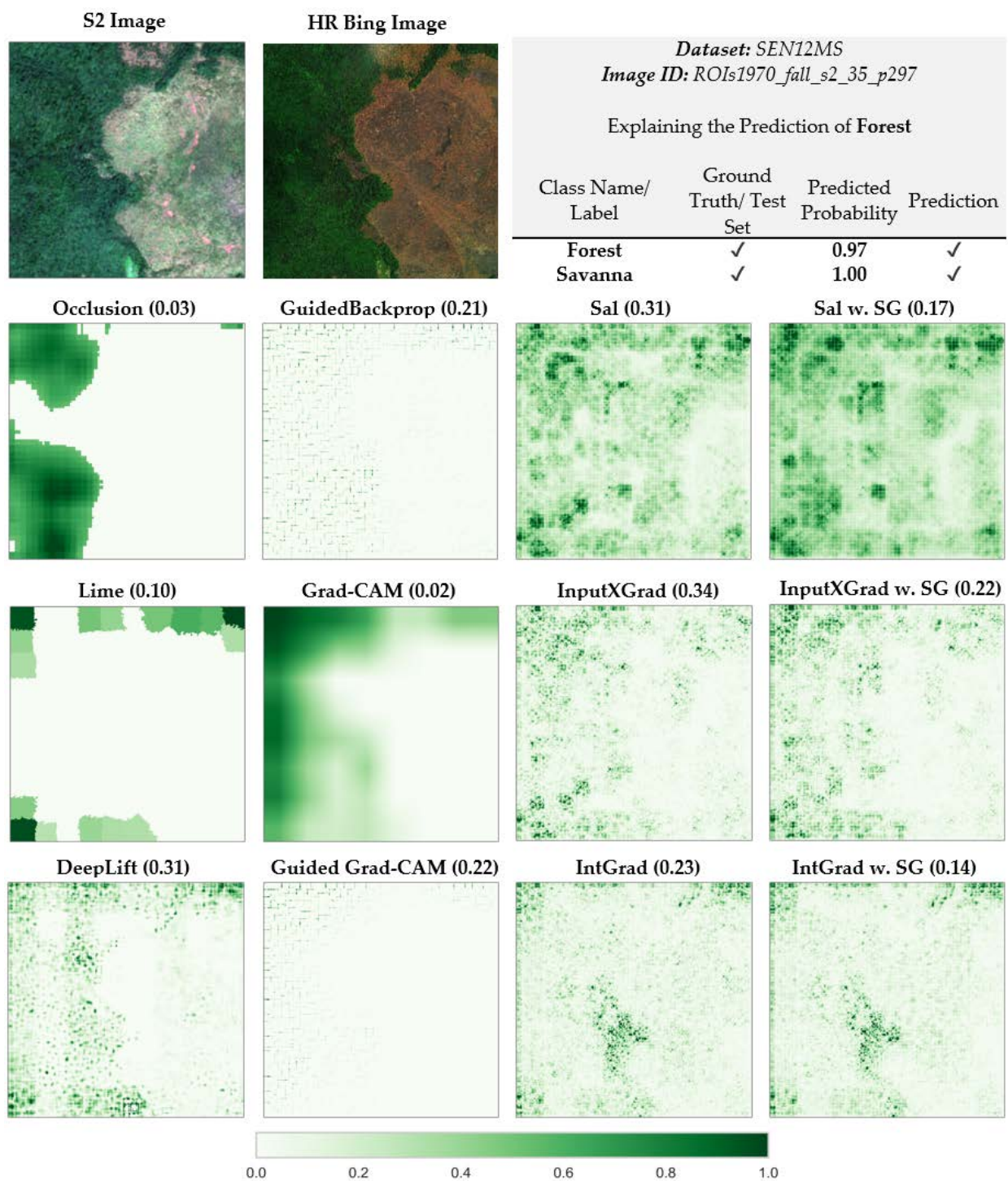

**Figure S5a.** Explaining the Predictions of DenseNet for *Forest* in SEN12MS (Image ID: ROIs1970_fall_s2_35_p297).

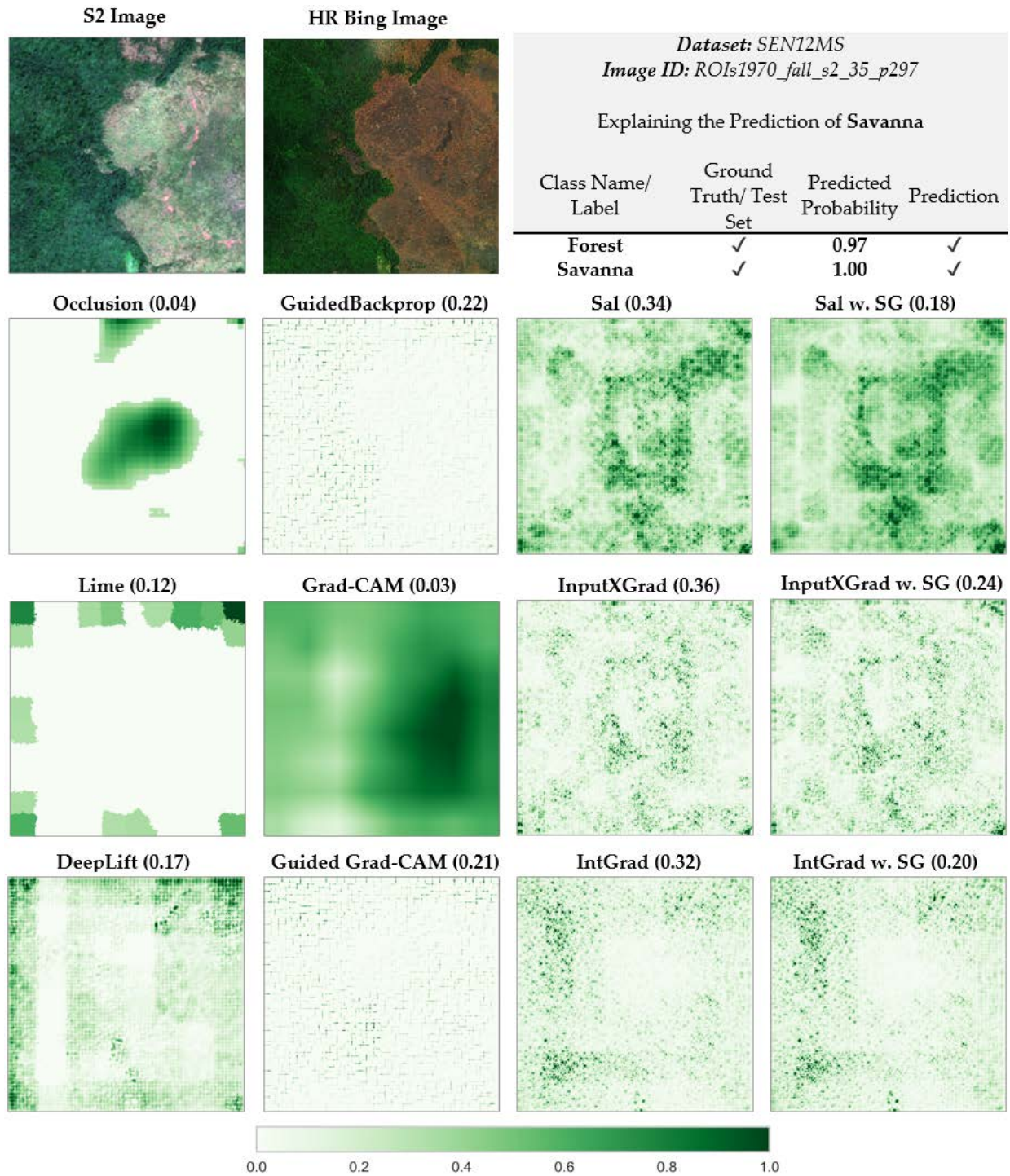

**Figure S5b.** Explaining the Predictions of DenseNet for *Savanna* in SEN12MS (Image ID: ROIs1970_fall_s2_35_p297).

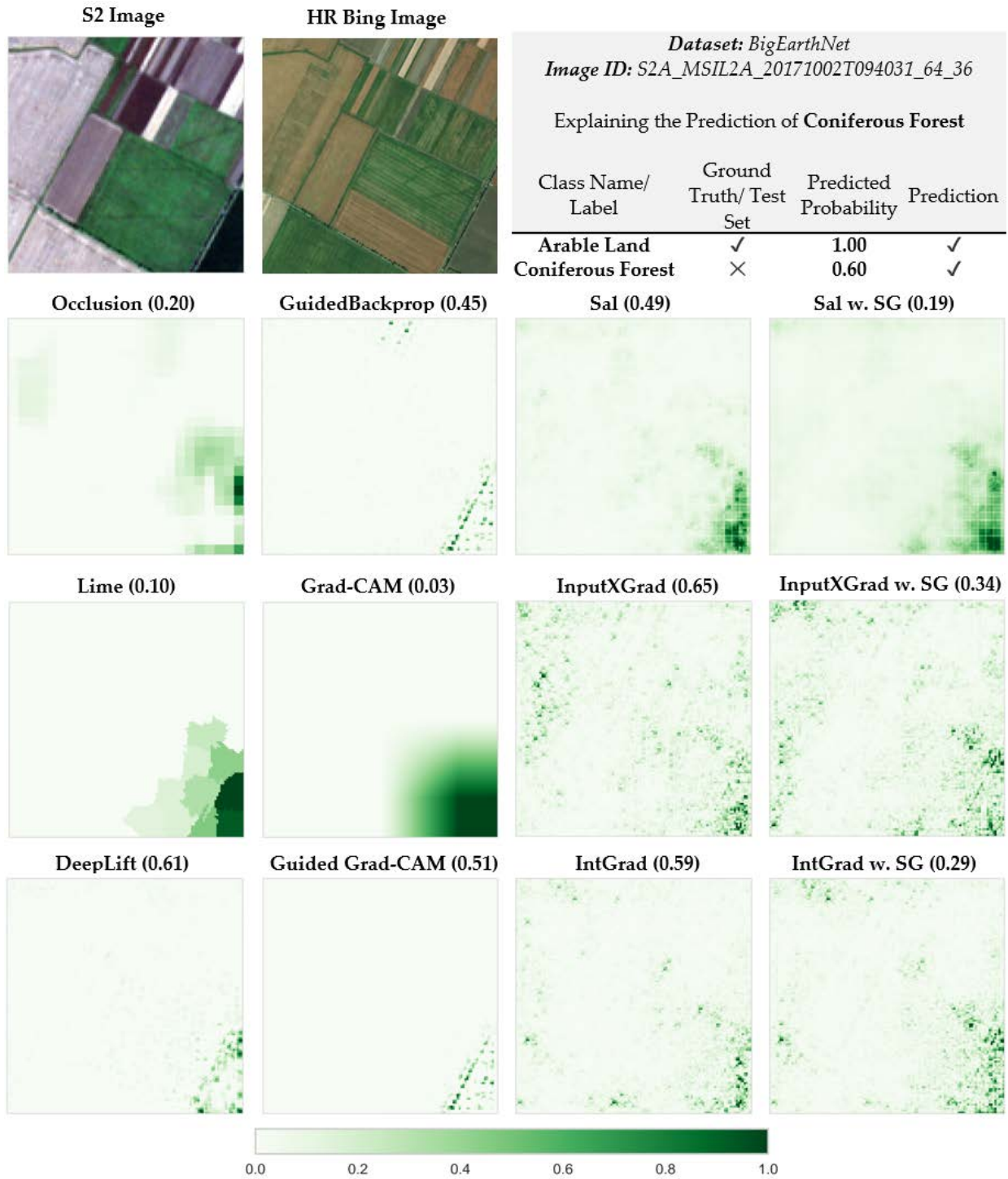

**Figure S6.** Explaining the Predictions of DenseNet for *Coniferous Forest* in BigEarthNet (Image ID: S2A_MSIL2A_20171002T094031_64_36).

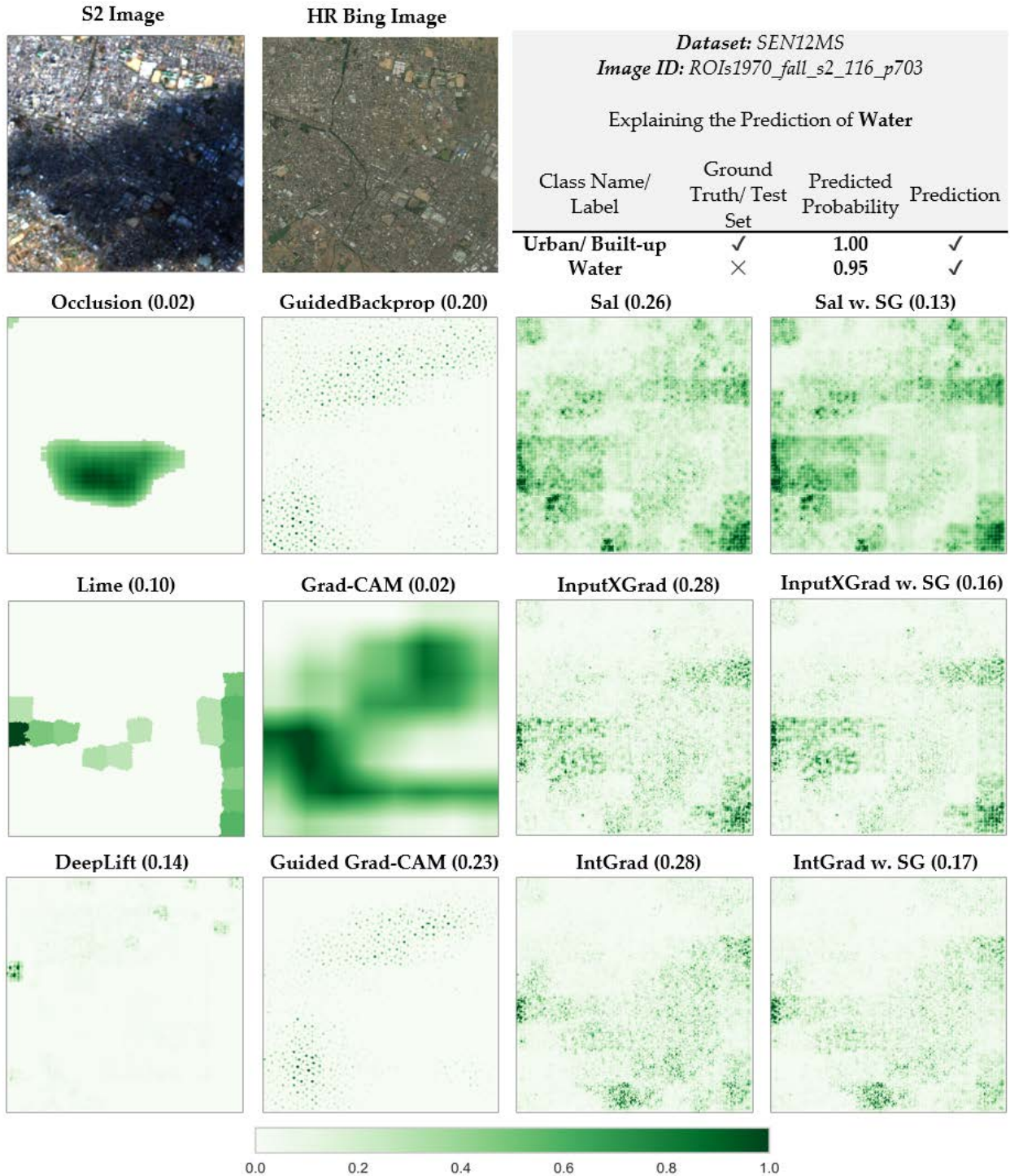

**Figure S7.** Explaining the Predictions of DenseNet for *Water* in SEN12MS (Image ID: ROIs1970_fall_s2_116_p703).

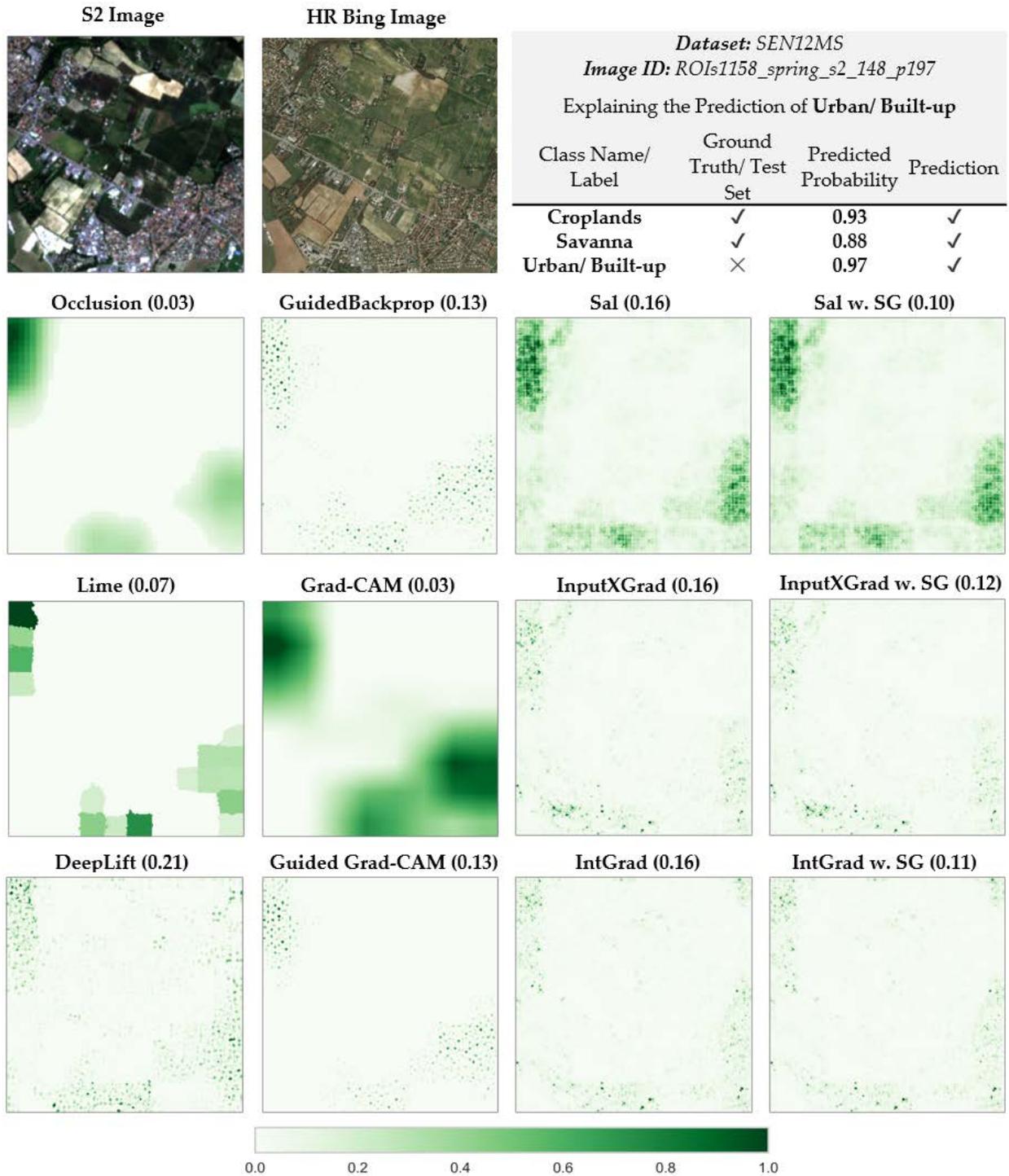

**Figure S8.** Explaining the Predictions of DenseNet for *Urban/ Built-up* in SEN12MS (Image ID: ROIs1158_spring_s2_148_p197).

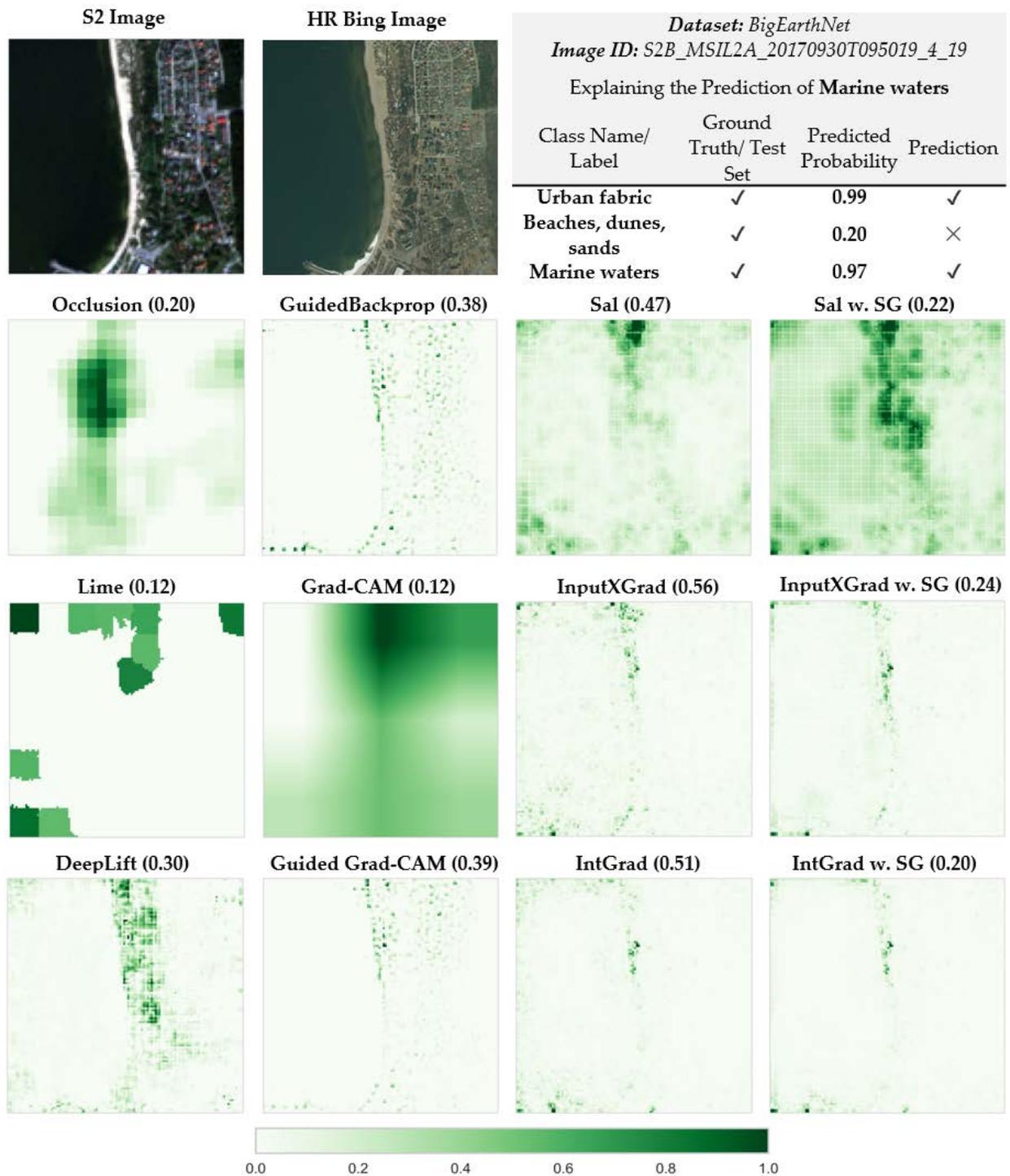

**Figure S9.** Explaining the Predictions of DenseNet for *Marine Waters* in BigEarthNet (Image ID: S2B_MSIL2A_20170930T095019_4_19).

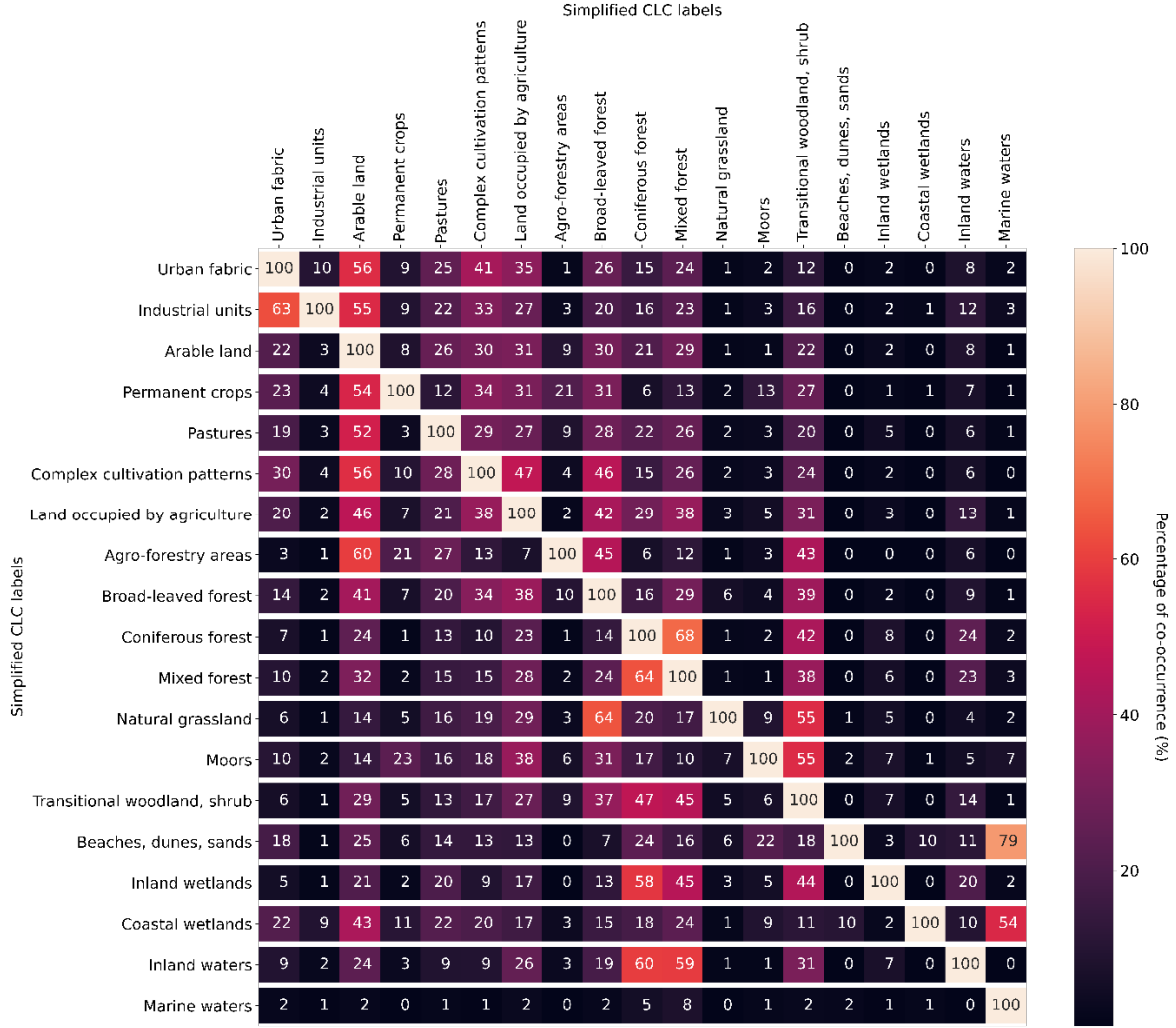

**Figure S10.** BigEarthNet training set co-occurrence

Table S1. DenseNet121 Per Class $F_1$ Scores on the BigEarthNet Dataset.

| Class | Mixed Forest | Natural Grassland | Moors, Heathland | Transitional Woodland-shrub | Beaches, Dunes, Sands | Inland Wetlands | Coastal Wetlands | Inland Waters | Marine Waters | Coniferous Forest |
|---|---|---|---|---|---|---|---|---|---|---|
| $F_1$-score (%) | 83.79 | 55.22 | 68.57 | 69.97 | **63.30** | 66.79 | 63.52 | 87.49 | 98.70 | 87.79 |
| Class | Urban Fabric | Industrial or Commercial Units | Arable Land | Permanent Crops | Pastures | Complex Cultivation Patterns | Land Occupied by Agriculture | Agro-forestry Areas | Broad-leaved Forest | Average |
| $F_1$-score (%) | 79.51 | 52.79 | 87.02 | 68.68 | 77.65 | 72.93 | 70.18 | 79.97 | 79.82 | 74.4 |

Table S2. DenseNet121 & ResNet50 Per Class $F_1$ Scores on the SEN12MS Dataset.

| Class | Forest | Shrubland | Savanna | Grassland | Wetlands | Croplands | Urban/Built-up | Snow/Ice | Barren | Water | Average |
|---|---|---|---|---|---|---|---|---|---|---|---|
| **DenseNet121** | **76.90** | **46.14** | **84.65** | 70.20 | 64.48 | 74.89 | **79.12** | 0 | 58.85 | 79.52 | 63.48 |
| **ResNet50** | 73.97 | 43.94 | 84.06 | **70.58** | **66.39** | **76.20** | 78.94 | 0 | **68.46** | **82.16** | **64.47** |

Table S3. Quantitative Metrics DenseNet121 & ResNet50 on the SEN12MS (lower scores indicate higher performance for all metrics).

| Method | Max-Sensitivity | | AUC-MoRF | | File Size (KB) | |
| --- | --- | --- | --- | --- | --- | --- |
| | DenseNet121 | ResNet50 | DenseNet121 | ResNet50 | DenseNet121 | ResNet50 |
| Sal | 0.27 | 0.35 | 36.92 | 33.96 | 23.35 | 19.57 |
| Sal w. SG | 0.14 | 0.16 | 36.32 | 33.30 | 22.54 | 18.94 |
| InputXGrad | 0.30 | 0.39 | 38.80 | 37.13 | 18.07 | 15.40 |
| InputXGrad w. SG | 0.16 | 0.21 | 38.67 | 37.31 | 17.68 | 14.88 |
| IntGrad | 0.26 | 0.30 | 37.40 | 36.69 | 17.91 | 16.43 |
| IntGrad w. SG | 0.13 | 0.16 | 37.64 | 36.79 | 16.72 | 15.26 |
| Guided Backprop | 0.23 | 0.27 | 34.67 | 36.97 | 22.68 | 10.90 |
| Grad-CAM | **0.03** | **0.05** | **23.62** | **25.70** | 6.31 | **6.45** |
| Guided Grad-CAM | 0.23 | 0.30 | 35.71 | 37.16 | 16.55 | 7.45 |
| DeepLift | 0.25 | 0.36 | 34.65 | 34.39 | 19.33 | 15.82 |
| Occlusion | **0.03** | 0.06 | 28.87 | 28.50 | **3.84** | 7.28 |
| Lime | 0.11 | 0.18 | 27.65 | 28.06 | 7.91 | 8.65 |